\documentclass[lettersize,journal]{IEEEtran}

\usepackage{booktabs, caption}
\usepackage{multirow, makecell}
\usepackage{xcolor}
\usepackage{subcaption}
\usepackage{fontawesome}

\usepackage{amsmath,amsfonts}
\usepackage{algorithmic}
\usepackage{algorithm}
\usepackage{array}
\usepackage[caption=false,font=normalsize,labelfont=sf,textfont=sf]{subfig}
\usepackage{textcomp}
\usepackage{stfloats}
\usepackage{url}
\usepackage{verbatim}
\usepackage{graphicx}
\usepackage{cite}

\usepackage{circledsteps}

\newtheorem{lemma}{Lemma}
\newtheorem{theorem}{Theorem}

\begin{document}

\title{Geometric Knowledge-Assisted Federated Dual Knowledge Distillation Approach Towards Remote Sensing Satellite Imagery}

\author{Luyao Zou, Fei Pan, Jueying Li, Yan Kyaw Tun,~\IEEEmembership{Senior Member,~IEEE,} Apurba Adhikary, Zhu Han, ~\IEEEmembership{Fellow,~IEEE,} and Hayoung Oh
\thanks{This work has been submitted to the IEEE for possible publication. Copyright may be transferred without notice, after which this version may no longer be accessible.}
\thanks{Luyao Zou and Hayoung Oh are with College of Computing and Informatics, Sungkyunkwan University, South Korea. (e-mail: zouluyao@skku.edu, hyoh79@skku.edu).}
\thanks{Fei Pan is with the Department of Computer Science and Engineering, University of Michigan, United States. (e-mail: feipanir@gmail.com)}
\thanks{Jueying Li is with Konkuk Aerospace Design-Airworthiness Institute, Konkuk University, Seoul, 05029, South Korea. (e-mail: dlrkdud@aliyun.com)}
\thanks{Yan Kyaw Tun is with the Department of Electronic Systems, Aalborg University, 2450 København SV , Denmark, (e-mail: {ykt}@es.aau.dk).}
\thanks{Apurba Adhikary is with the Department of Information and Communication Engineering, Noakhali Science and Technology University, Noakhali-3814, Bangladesh (e-mail: apurba@khu.ac.kr).}
\thanks{Zhu Han is with  the Department of Electrical and Computer Engineering at the University of Houston, Houston, TX 77004 USA (e-mail: hanzhu22@gmail.com).}
\thanks{Hayoung Oh is the corresponding author.}
}

\markboth{Journal of \LaTeX\ Class Files,~Vol.~14, No.~8, August~2021}%
{Shell \MakeLowercase{\textit{et al.}}: A Sample Article Using IEEEtran.cls for IEEE Journals}


\maketitle

\begin{abstract}
Federated learning (FL) has recently become a promising solution for analyzing remote sensing satellite imagery (RSSI). However, the large scale and inherent data heterogeneity of images collected from multiple satellites, where the local data distribution of each satellite differs from the global one, present significant challenges to effective model training. To address this issue, we propose a Geometric Knowledge-Guided Federated Dual Knowledge Distillation (GK-FedDKD) framework for RSSI analysis. In our approach, each local client first distills a teacher encoder (TE) from multiple student encoders (SEs) trained with unlabeled augmented data. The TE is then connected with a shared classifier to form a teacher network (TN) that supervises the training of a new student network (SN). The intermediate representations of the TN are used to compute local covariance matrices, which are aggregated at the server to generate global geometric knowledge (GGK). This GGK is subsequently employed for local embedding augmentation to further guide SN training. We also design a novel loss function and a multi-prototype generation pipeline to stabilize the training process. Evaluation over multiple datasets showcases that the proposed GK-FedDKD approach is superior to the considered state-of-the-art baselines, e.g., the proposed approach with the Swin-T backbone surpasses previous SOTA approaches by an average $68.89\%$ on the EuroSAT dataset.
\end{abstract}


\section{Introduction}
\label{sec:intro}
\IEEEPARstart{S}{atellite} imagery, an essential product of remote sensing \cite{liu2025effective}, is able to track the Earth surface's dynamic, which can offer valuable information to assist decision-makers in multiple application fields \cite{capliez2023temporal}. Besides, these data can also provide solutions to global challenges such as biodiversity loss and climate change~\cite{mall2023change}. Moreover, satellite imagery has become increasingly important in a wide range of realms. Particularly, in recent years, satellite imagery (especially optical satellite images) has been widely applied in diverse fields, such as land cover classification and agriculture monitoring \cite{zou2024diffcr}.  


\begin{figure} [t!]
	\centering
	\includegraphics[scale = 0.37]{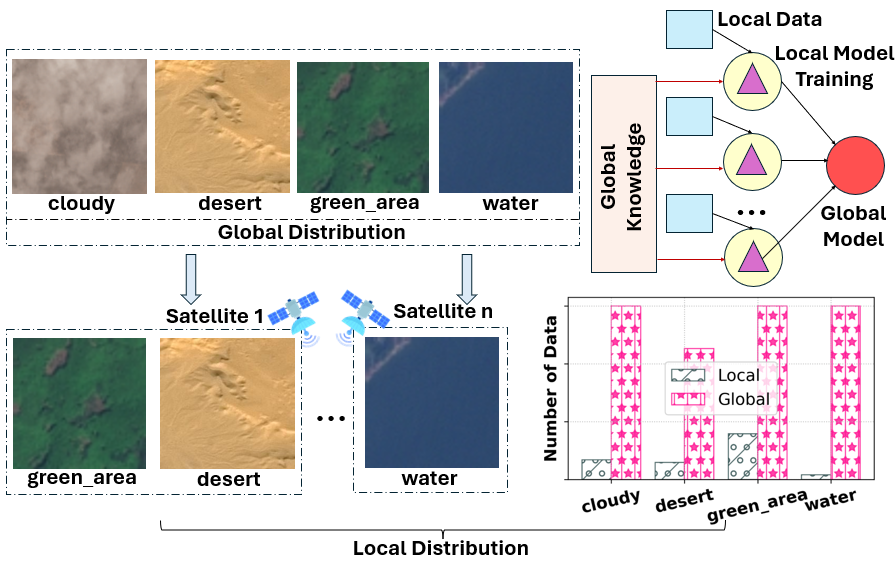}
	\captionsetup{font=small}
	\caption{The discrepancy between local \& global distribution for satellite imagery. Each satellite may have different categories and different amounts of data. Images are derived from the \textit{Satellite Image Classification (SIC) Dataset} \cite{satelliteClassficationDataset}.}
	\label{fig_dicrepancy_local——global_distribution}
\end{figure}

\par
Nevertheless, remote sensing satellite imagery (RSSI) collected from multiple satellites is often massive in scale and exhibits non-independent and identically distributed (non-IID) characteristics, commonly referred to as data heterogeneity \cite{li2021federated}. This heterogeneity arises from discrepancies between local and global data distributions \cite{ma2025geometric}, as illustrated in Fig. \ref{fig_dicrepancy_local——global_distribution}. Specifically, each satellite may capture only a subset of the categories present in the global dataset, and the number of samples per category can vary significantly across satellites. To be more specific, as demonstrated in Fig. \ref{fig_dicrepancy_local——global_distribution}, the global distribution comprises four categories: $\{\texttt{cloudy, desert, green\_area, water}\}$, while each satellite (local distribution) may only contain a portion of the classes. For instance, "Satellite $1$" only acquires data belonging to two classes: $\{\texttt{green\_area, desert}\}$, and "Satellite $2$" merely has data in the $\texttt{water}$ category. In addition, although some satellites may own the image data of the same category, the number of images may differ. These imbalances raise the key challenges: 
\begin{itemize}
\item How to mitigate the data distribution gaps among different satellites?
\item How to mitigate the gaps between local and global data distributions while learning from RSSI?
\end{itemize}


\par
To address these challenges, federated learning (FL) has emerged as a promising distributed paradigm that enables multiple satellites to collaboratively train models while preserving data privacy. However, traditional FL methods can potentially suffer from performance degradation under data heterogeneity \cite{zou2025towards}. To mitigate this issue, various studies \cite{zou2025towards, ding2024global, seo2024relaxed, hu2024fedmut, zhang2025lcfed, tan2022fedproto, tamirisa2024fedselect, zou2025prototype, duan2023fedac} have enhanced FL through sever-side strategies (e.g., improved server-side aggregation) and/or client-side strategies (e.g., client-side knowledge transfer strategies). Furthermore, recent works \cite{ma2025geometric, sun2024fedcpd, ye2023feddisco, morafah2024stable, xiao2025federated, yan2025simple, zhang2023data} have begun addressing the discrepancy. Nevertheless, among the studies of \cite{tan2022fedproto, zhang2025lcfed, ding2024global, tamirisa2024fedselect, seo2024relaxed, hu2024fedmut, zou2025towards, zou2025prototype, duan2023fedac, ma2025geometric, xiao2025federated, morafah2024stable, zhang2023data, yan2025simple, sun2024fedcpd, ye2023feddisco}, none of them pay attention to the joint consideration of model aggregation, prototype aggregation, local client model construction, and the data distribution
discrepancy mitigation by global knowledge. To this end, in this work, we are dedicated to designing an approach that covers these aspects. The main contribution of this work is summarized as follows: 
\begin{itemize}
\item In this work, we propose a method called the geometric knowledge-guided federated dual knowledge distillation (GK-FedDKD) approach, which contains data augmentation, dual knowledge distillation, multi-prototype learning, model aggregation, and global geometric knowledge extraction (GKE) to effectively handle non-IID satellite imagery. 
\item We develop a dual-knowledge distillation (KD) approach to construct the teacher encoder (TE) and the entire local model for each client (i.e., a satellite), respectively. Specifically, the dual KD includes 1) KD with unlabeled data and 2) KD with labeled original data and a shared classifier. 
\item Particularly, to acquire the TE, the first KD is used by concurrently training several student encoders (SEs) on the augmented data with no labels. To train the entire local model, the second KD is utilized, where we attempt to regulate local training by additionally augmenting the local embedding based on global geometric knowledge obtained by the server, which will be the input of the classifier.
\item We also design a linear layer-based module for each client, where we use an extra linear layer to map the output dimension space of the student encoder to the dimension space of labels in one-hot vector format to create a loss function.
\item For the server, we design the server to take multiple roles simultaneously: model aggregation, multi-prototype aggregation, and GKE (i.e., global geometric shape generation and global vector production).
\item Experimental results convince the effectiveness of the proposed method in handling the satellite image compared to the state-of-the-art methods such as FedExP \cite{jhunjhunwala2023fedexp}, FedAU \cite{wang2023lightweight}, and FedAS \cite{yang2024fedas}. For instance, when using the SAT4 dataset and the ResNet10 backbone, the proposed GK-FedDKD approach improves accuracy by $1.76\%\sim75.34\%$ when the Dirichlet distribution parameter is 0.9. The performance enhancement can also be observed in other cases (e.g., across various datasets).
\end{itemize}
\par
The remainder of the paper is organized as follows. In Section \ref{sec_2_related_works} we offer several related works. The details of the proposed GK-FedDKD method are provided in Section \ref{sec_3_methodology}, while a comparison experiment is conducted in Section \ref{sec_4_experiments}. Finally, in Section \ref{sec_5_conclusion}, we conclude this article.

\section{Related Works}
\label{sec_2_related_works}
Data heterogeneity issues and data distribution disparities are the two main concerns of this work. Hence, this section will elaborate on the previous literature based on these two aspects. In addition, the differences between this article and the related works are also provided.

\subsection{FL towards Data Heterogeneity}
\label{FL_Towards_Data_Heterogeneity}
To enhance the performance of conventional FL in addressing the heterogeneous data, prior works such as \cite{zou2025towards, ding2024global, seo2024relaxed, hu2024fedmut, zhang2025lcfed, tan2022fedproto, tamirisa2024fedselect,  zou2025prototype, duan2023fedac} have been proposed. Before delving into those studies, it deserves to introduce FedAvg \cite{mcmahan2017communication} that is a seminal method to FL. In FedAvg, a global model can be obtained by taking a weighted average of the local updated models. Based on FedAvg, \cite{zou2025towards, ding2024global, seo2024relaxed, hu2024fedmut} were proposed. Specifically, \cite{zou2025towards} proposed OSC-FSKD approach, a clustered FL method, to process satellite images with non-IID features, where FedAvg will be performed in each round to produce the cluster-level global model, and it will be executed across all the clients in the final round to generate a global model. However, \cite{zou2025towards} does not take into account multi-prototype learning and the differences between global and local distributions, whereas those aspects are involved in this work. In \cite{ding2024global}, to adapt to data heterogeneity scenarios, FedAKD was proposed, which is a generalized FL approach that allows the coexistence of both the personalized models and the global model. Different from \cite{ding2024global}, we consider a global model and multiple prototypes in this work. In \cite{seo2024relaxed}, to improve the performance of FL in processing heterogeneous data, FedRCL was proposed, in which a relaxed contrastive learning loss was presented. Nevertheless, multi-prototype learning is not considered by \cite{seo2024relaxed}, whereas that is one of the considerations in this paper. In \cite{hu2024fedmut}, FedMut was derived from FedAvg, where the global model is used to produce mutated models via a general mutation strategy during local training. In contrast with \cite{hu2024fedmut}, rather than generating mutated models, we follow FedAvg to send the global model to the local client for the next round of training, and we also design a linear layer-based module to facilitate the local model training. Like \cite{zou2025towards}, clustered FL was also considered by \cite{zhang2025lcfed}. Specifically, in \cite{zhang2025lcfed}, instead of a weighted average
manner (i.e., FedAvg), the server produces multiple cluster center models by simply averaging the local models belonging to the same cluster, and also, the server generates a global embedding by aggregating the shallow embedding of each client. \cite{tan2022fedproto} proposed a novel aggregation scheme, in which class prototypes are aggregated in the server. Nonetheless, both \cite{zhang2025lcfed} and \cite{tan2022fedproto} do not consider multi-prototype learning and KD, which are the focus of this work. In \cite{tamirisa2024fedselect}, partial model personalization was considered to design the FL-based approach, called FEDSELECT. In \cite{zou2025prototype}, Pro-FedKD was proposed to deal with satellite image data, which is constructed based on the FL scheme, self-KD, and federated prototype learning (FedProto). In \cite{duan2023fedac}, FedAC was proposed to solve the data heterogeneity problem, where a model initialization algorithm and an alternating contrastive training algorithm were proposed. Unlike prior works \cite{tamirisa2024fedselect, zou2025prototype, duan2023fedac}, this paper additionally considers the inconsistency between the local and global distributions. 

\begin{figure*} [t!]
	\centering
	\includegraphics[scale = 0.68]{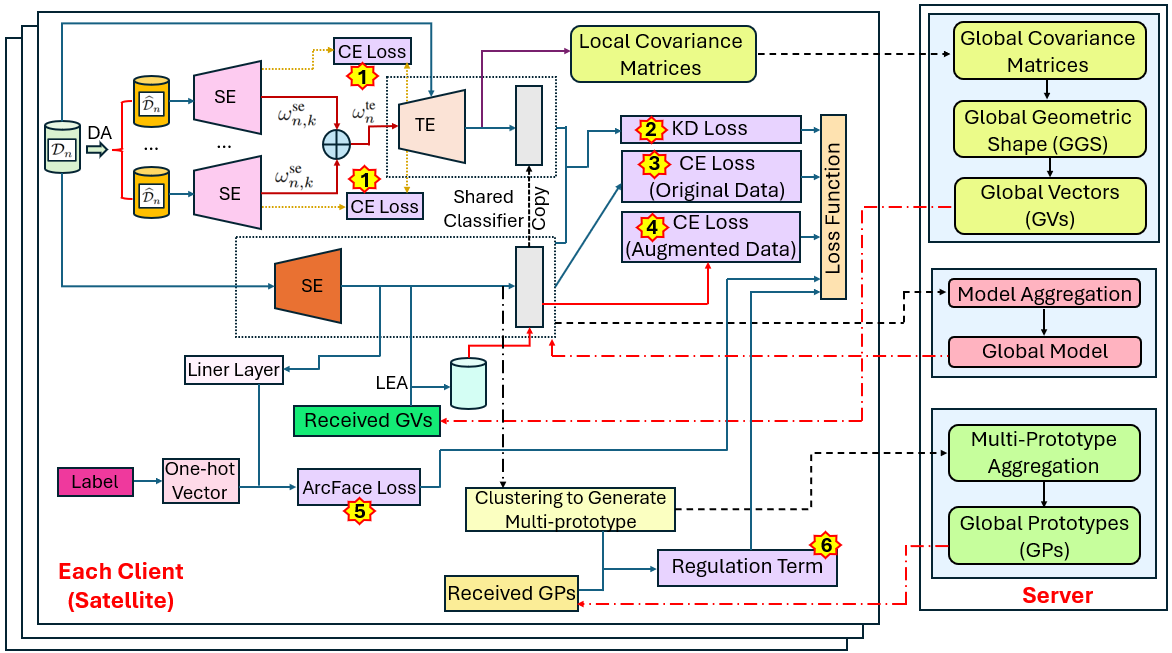}
	\captionsetup{font=small}
	\caption{Architecture of the proposed GK-FedDKD method, which is formed by multiple clients (i.e., satellites) and a server. This architecture showcases the specific implementation of Fig. \ref{fig_solution}.}
	\label{fig_architecture}
    \vspace{-0.3cm}
\end{figure*}

\subsection{Bridging Data Distribution Inconsistency}
\label{DA_Data_Heterogeneity_Improvement}
Data heterogeneity arises from the disparity of data held by the clients \cite{tang2022virtual}. Therefore, it is valuable to summarize several prior studies that consider data distribution discrepancy (DDD). DDD has been investigated in several works, such as \cite{ma2025geometric, sun2024fedcpd, ye2023feddisco, morafah2024stable, xiao2025federated, yan2025simple, zhang2023data}. Specifically, in \cite{ma2025geometric}, a geometry-guided method was proposed to generate data. Although the data-generation approach of this paper is motivated by \cite{ma2025geometric}, the global vector generation method in this work differs from that in \cite{ma2025geometric}. As for both \cite{sun2024fedcpd} and \cite{ye2023feddisco}, they accounted for the discrepancy between the local and global sides and proposed discrepancy-based aggregation weights to guide the generation of the global model. In particular, in \cite{sun2024fedcpd}, a method called FedCPD was proposed, in which local category inconsistency is computed for each client by comparing the local and global data distributions. The server will collect the calculated local category inconsistency to obtain the aggregation weights. In \cite{ye2023feddisco}, a method named FedDisco was presented, in which model aggregation is conducted using a discrepancy-aware aggregation weight that depends on the discrepancy between the local and global category distributions. Unlike \cite{sun2024fedcpd} and \cite{ye2023feddisco}, we consider using geometric knowledge to augment local data and mitigate the discrepancy between local and global data distributions. To confront the data distribution inconsistency problem, data augmentation has also been introduced in FL by \cite{morafah2024stable, xiao2025federated, yan2025simple, zhang2023data}. In particular, in \cite{morafah2024stable}, Gen-FedSD was proposed, which involved pre-trained stable diffusion-based data augmentation. In \cite{xiao2025federated}, the proposed FedEqGAN helps balance differences in data distribution across clients by generating synthetic data that preserves similar statistical properties. In \cite{yan2025simple}, an input-level data augmentation method, FedRDN, was proposed. In this method, data augmentation is performed by manipulating the channel-wise data statistics of multiple clients during the training stage. In \cite{zhang2023data}, FedM-UNE was presented based on data augmentation, employing MixUp (a classic data augmentation method) in federated settings and avoiding raw data transfer. Unlike \cite{morafah2024stable, xiao2025federated, yan2025simple, zhang2023data}, we consider using the geometric knowledge to lead the data augmentation process. In addition, dual KD, multiple prototype learning, and the linear-layer-based module considered in this work are not included in \cite{morafah2024stable, xiao2025federated, yan2025simple, zhang2023data}.

\section{Methodology}
\label{sec_3_methodology}

\subsection{Overview}



\par
Fig. \ref{fig_architecture} demonstrates the architecture of the proposed GK-FedDKD approach for remote sensing satellite imagery. Specifically, as shown in Fig. \ref{fig_architecture}, the proposed GK-FedDKD approach consists of two major components: the client (i.e., satellite) side and the server side. Each client contains: 1) KD-based TE generation by using multiple SEs with unlabeled augmented data, 2) the generated TE will be leveraged to compute the local covariance matrices (LCM), meanwhile, the TE connects with a shared classifier that will be used to guide a new student network training, 3) the output of SE will be augmented by using the global vectors (GVs) \cite{ma2025geometric} generated by the server, which is regarded as the input of the classifier to regulate the local training, 4) a linear layer module to form a local loss, and 5) the multi-prototype generation policy. Each client interacts with the server by transferring the LCM, the local model, and the multi-prototype. With respect to the server, it contacts each client by separately disseminating the generated GVs, the global model, and the global prototypes. 

\begin{figure*} [t!]
	\centering
	\includegraphics[scale = 0.6]{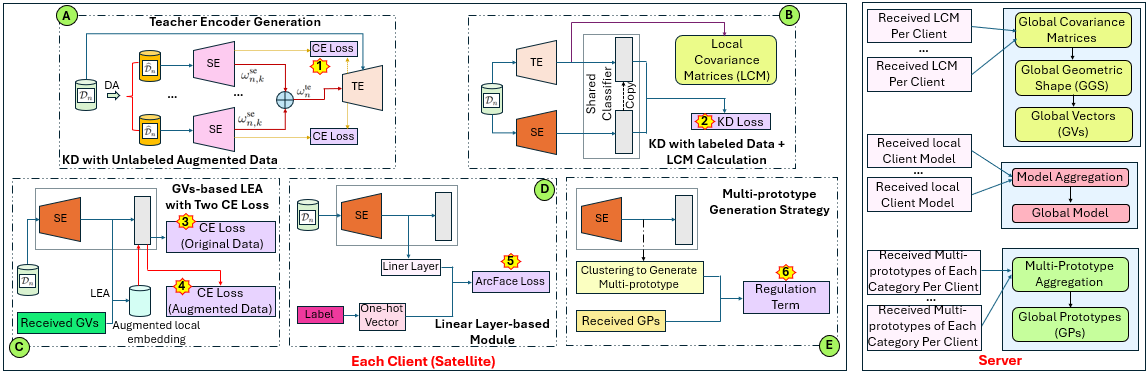}
	\captionsetup{font=small}
	\caption{The overall framework of the proposed GK-FedDKD method includes: client \& server parts. The client part contains: \Circled{A} 
    KD with unlabeled augmented data for TE generation, \Circled{B} KD with labeled data and local covariance matrices calculation, \Circled{C} GVs-guided local embedding augmentation (LEA) and local learning regulation, \Circled{D} a linear layer-based module, and \Circled{E} multi-prototype (MP) generation strategy. The server part plays the role of 1) global covariance matrices generation, 2) model aggregation, and 3) MP aggregation.}
	\label{fig_solution}
    \vspace{-0.3cm}
\end{figure*}

\par
For ease of understanding, we provide the overall framework in Fig. \ref{fig_solution} to illustrate the detailed components of the proposed GK-FedDKD approach. Concretely, each satellite client includes five key modules, listed as follows: 
\begin{itemize}
\item Module \Circled{A}: generation of a TE via KD using unlabeled augmented data;
\item Module \Circled{B}: KD with labeled data and computation of LCM;
\item Module \Circled{C}: data augmentation (DA) guided by GVs received from the server;
\item Module \Circled{D}: a linear-layer module for supervised learning;
\item Module \Circled{E}: a multi-prototype generation \cite{zhang2023dual} module.
\end{itemize}
Detailed descriptions of these components are provided in the following sections. It is noteworthy that we also provide the entire architecture of the proposed GK-FedDKD approach in Fig. \ref{fig_architecture}, to demonstrate how each module works.

\subsection{Teacher Encoder Generation via KD with Unlabeled Augmented Data}
\label{Teacher_Encoder_Generation_KD_No_Labels}
To construct the TE, we employ KD with the utilization of the unlabeled augmented data. Specifically, multiple student encoders are utilized to distill and generate the TE. This design is inspired by~\cite{caron2021emerging}, yet our approach differs in several key aspects, shown as follows: 
\begin{itemize}
\item Unlike~\cite{caron2021emerging}, which adopts both global and low-resolution local views as inputs, our method feeds the original data into the TE, while the SEs receive augmented variants of the same data produced by standard data augmentation techniques such as rotation, Gaussian noise, flipping, and salt-and-pepper perturbations.
\item Instead of updating the TE via an exponential moving average of SE parameters, we employ a simple linear combination strategy (see Part A of Fig. \ref{fig_solution}).
\end{itemize}
The model of TE, $\omega_n^\textrm{te}$, for client $n$ is obtained by
\begin{subequations}\label{Opt_1}
	\begin{align}
		\omega_n^\textrm{te} = \sum_{k=1}^{K}{\lambda_k\omega_{n, k}^\textrm{se}}, \tag{1} 
	\end{align}	
\end{subequations}
where $\lambda_k$ is the hyperparameter, $K$ denotes the total number of SEs, and $\omega_{n, k}^\textrm{se}$ represents the model of the $k^\textrm{th}$ SE on the client $n$. To update $\omega_{n, k}^\textrm{se}$, inspired by \cite{caron2021emerging}, we consider training SE to match the output of the TE. At this stage, cross-entropy (CE) loss is adopted. Let $\mathcal{N}=\{1, 2, \ldots, N\}$ denote a set of satellites, and let $\mathcal{D}_n = \{(x_n^\textrm{1}, y_n^\textrm{1}), \ldots, (x_n^\textrm{m}, y_n^\textrm{m})\}$ and $\widehat{\mathcal{D}}_n = \{(\widehat{x}_n^\textrm{1}, y_n^\textrm{1}), \ldots, (\widehat{x}_n^\textrm{m}, y_n^\textrm{m})\}$ represent the local data owned by each satellite $n$ and the corresponding augmented data, respectively. Then, we define CE loss as follows \cite{caron2021emerging}:
\begin{subequations}\label{Opt_2}
	\begin{align}
		\mathcal{H}_n^\textrm{CE}\Big(p_n^\textrm{te}(x_n^\textrm{m}), p_n^\textrm{se}(\widehat{x}_n^\textrm{m})\Big) = -p_n^\textrm{te}(x_n^\textrm{m}) \log p_n^\textrm{se}(\widehat{x}_n^\textrm{m}), \tag{2} 
	\end{align}	
\end{subequations}
where $p_n^\textrm{te}(x_n^\textrm{m})$ and $p_n^\textrm{se}(\widehat{x}_n^\textrm{m})$ denote the output probability of TE and SE obtained by the softmax function, respectively.
 
\subsection{Local Client Construction via KD}
\label{local_client_construction_KD}
To build the local client model, another KD is applied (Part B of Fig. \ref{fig_solution}). Concretely, we let both the TE generated in Section \ref{Teacher_Encoder_Generation_KD_No_Labels} and a new SE connect with a \textbf{shared classifier} to build the teacher network (TN) and the student network (SN), respectively. It is worth noting that the shared classifier in this work means the parameters of the classifier of the TN and SN are the same. During training, we only train SN. For TN, we freeze the TE part but update the classifier by copying from SN directly. The original data is employed as the input of TN and SN. The loss for KD is defined as follows \cite{zou2025towards}:
\begin{subequations}\label{Opt_3}
	\begin{align}
		 \mathcal{L}_n^\textrm{KD} =  \frac{1}{|\mathcal{D}_n|}\sum_{m=1}^{|\mathcal{D}_n|}\tau^2KL\Big(\varphi_n(x_n^\textrm{m}), \widehat{\varphi_n}(x_n^\textrm{m})\Big), \tag{3} 
	\end{align}	
\end{subequations}
where $\tau$ indicates the temperature. $KL(\cdot, \cdot)$ is adopted to represent the Kullback-Leibler (KL) divergence. $\varphi_n(x_n^\textrm{m})$ and $\widehat{\varphi_n}(x_n^\textrm{m})$ represent the soft probabilities of TN and SN, respectively, which can be acquired by employing \textit{Softmax with Temperature scaling} \cite{zou2024cyber}. $|\mathcal{D}_n|$ denotes the number of data owned by client $n$.



\subsection{Local Covariance Matrices and Local Embedding Augmentation}
\label{local_covariance_matrices}
Apart from the network structure (Section \ref{local_client_construction_KD}), local covariance matrices (LCM) and local embedding augmentation (LEA) are also considered in the sequel.
\par
\textbf{Local Covariance Matrices.} \quad As LCM is essential for getting the geometric shape as per \cite{ma2025geometric}, LCM is considered, as shown in Part B of Fig. \ref{fig_solution}. Particularly, LCM of each satellite $n$ per class $c$ will be computed by using the output of TE. Following \cite{ma2025geometric}, for each class $c$, we calculate the local covariance matrix (denoted as $\varrho_n^\textrm{c}$) as follows:
\begin{subequations}\label{Opt_4}
	\begin{align}
		 \varrho_n^\textrm{c} = \frac{1}{|\mathcal{D}_{n, c}|}\sum_{m=1}^{|\mathcal{D}_{n, c}|}(x_n^\textrm{m,c}-\mu_n^\textrm{c})(x_n^\textrm{m,c}-\mu_n^\textrm{c})^\top, \tag{4} 
	\end{align}	
\end{subequations}
where $x_n^\textrm{m,c}$ denotes the $m^\textrm{th}$ data sample of client $n$ belonging to class $c$. $\mathcal{D}_{n, c}$ is the subset of $\mathcal{D}_{n}$, which contains the data belonging to class $c$ in client $n$. $|\mathcal{D}_{n, c}|$ represents the overall number of the data in client $n$ falling within category $c$. Besides, $\mu_n^\textrm{c}$ indicates the mean of the data belonging to class $c$ of client $n$, which is calculated by $\mu_n^\textrm{c}=\frac{1}{|\mathcal{D}_{n, c}|}\sum_{m=1}^{|\mathcal{D}_{n, c}|}x_n^\textrm{m,c}$. The obtained $\varrho_n^\textrm{c}$ and $\mu_n^\textrm{c}$ will be submitted to the server to acquire the global geometric shape (GGS), which will be further utilized to obtain the global vector per class (denoted as $\Omega_c$) on the server. The detailed process of getting $\Omega_c$ will be given in Section \ref{global_design}. 

\textbf{Local Embedding Augmentation.} \quad In this work, we consider letting the clients receive the global vectors from the server to augment the local embedding, where the augmented local embedding will be used as the input of the classifier to form a cross-entropy (CE) loss, as presented in Part C of Fig. \ref{fig_solution}. Particularly, the local embedding in this work is the output of SE. Given an input data $x_n^\textrm{m,c}$, the augmented local embedding, $\Upsilon_n^\textrm{m,c}$, is defined as
\begin{subequations}\label{Opt_5}
	\begin{align}
		\Upsilon_n^\textrm{m,c} = f_n(x_n^\textrm{m,c}) + \Omega_c. \tag{5} 
	\end{align}	
\end{subequations}
Here, $f_n(\cdot)$ is SE's embedding function for client $n$. As per \cite{sun2016max}, we give the CE loss for $\Upsilon_n^\textrm{m,c}$ as
\begin{subequations}\label{Opt_6}
	\begin{align}
		\mathcal{L}_n^\textrm{CEA} = \sum_{m=1}^{|\mathcal{D}_n|}l_{m}^\textrm{ce}, \tag{6} 
	\end{align}	
\end{subequations}
where $l_{m}^\textrm{ce} = -\sum_{c}q(c|\Upsilon_n^\textrm{m,c})\log p(c|\Upsilon_n^\textrm{m,c})$ \cite{wang2019symmetric}. Here, $q(c|\Upsilon_n^\textrm{m,c})$ denotes the ground-truth distribution over the corresponding labels of $\Upsilon_n^\textrm{m,c}$, while $p(c|\Upsilon_n^\textrm{m,c})$ is the probability of each category computed by the classifier for the given $\Upsilon_n^\textrm{m,c}$. For convenience, we refer to this process as \textit{global information alignment (GIA)}.

\subsection{Linear Layer-based Module}
To further boost performance, we incorporate an additional linear-layer module (Part D of Fig. \ref{fig_solution}), where cosine similarity is used to formulate the loss. Given the strong effectiveness of ArcFace loss \cite{deng2019arcface} in ameliorating supervised learning for cosine similarity\cite{zhang2024upload}, we design a loss function based on ArcFace loss. Different from \cite{deng2019arcface,zhang2024upload}, we use the SE output $f(x_n^\textrm{m})$ and its corresponding one-hot label to gain the angle. Because the dimensionality of $f(x_n^\textrm{m})$ may differ from that of the one-hot label, we introduce a linear layer that projects $f(x_n^\textrm{m})$ into a space to match the label dimension. Therefore, the designed loss (based on ArcFace loss) is formulated by
\begin{subequations}\label{Opt_7}
	\begin{align}
		\mathcal{L}_n^\textrm{AF} = -\frac{1}{|\mathcal{D}_n|}\sum_{n=1}^{|\mathcal{D}_n|}\log \frac{e^{\hat{s}(\cos{(\theta_y + h)})}}{e^{\hat{s}(\cos{(\theta_y + h)})}+\sum_{c=1, c\neq y}^{|\mathcal{C}_n|}e^{\hat{s}\cos{\theta_c}}}, \tag{7}
	\end{align}
\end{subequations} 
where $\theta_y$ denotes the angle between the output of the added linear layer and the label in one-hot vector format. $\hat{s}$ denotes a re-scale hyperparameter, while $h$ represents an additive angular margin penalty \cite{deng2019arcface}. $\mathcal{C}_n = \{1, 2, \ldots, C_n\}$ is the set of classes that are owned by client $n$, while $|\mathcal{C}_n|$ represents the number of categories of client $n$. 

\subsection{Multi-prototype Generation Policy}
\label{multi_prototype_generation_per_student_network}
As per \cite{tan2022fedproto}, the prototype-based mechanism can be applied to FL to solve the heterogeneous issue. In \cite{tan2022fedproto}, the average operation over the local embeddings belonging to the same class is employed to obtain the prototype of each class. However, such a manner will result in the omission of some useful information \cite{zhang2021self}. To grasp the feature information of the samples more comprehensively, according to \cite{zhang2023dual}, a multi-prototype generation policy (MPGP) can be introduced. Consequently, we apply MPGP to obtain the solution, as depicted in Part E of Fig. \ref{fig_solution}. For simplicity, following \cite{le2023fedmp, deuschel2021multi}, K-Means is adopted on the local embeddings of each class (i.e., the output of SN with original data), where the centroids per class are the prototypes. The corresponding mathematical representation can be defined as follows:
\begin{subequations}\label{Opt_8}
		\begin{align}
		\hspace{-0.1cm}	\{P_{n, c}^i\}_{i=1}^{|\mathcal{P}_{n, c}|} \xleftarrow{\textrm{centroids}}	KMeans(\{f_n(x_n^\textrm{m};y_n^\textrm{m}=c)\}_{m=1}^{\left| \mathcal{D}_{n, c} \right|}),  \tag{8} 
		\end{align}	
\end{subequations}
where $P_{n, c}^{i}$ represents the $i^\textrm{th}$ local prototype of the category $c$ for the client (satellite) $n$. $\mathcal{P}_{n, c} = \{P_{n, c}^1, P_{n, c}^2, \ldots, P_{n, c}^i\}$ is a set of local prototypes for category $c$ of client $n$. $|\mathcal{P}_{n, c}|$ is leveraged to denote the number of local prototypes of the same category $c$ for client $n$. Following \cite{tan2022fedproto}, the local prototypes of each class will be disseminated to the server to perform the aggregation process to generate global prototypes, denoted as $\mathcal{P} = \{P_1, P_2, \ldots, P_c\}$ (given in Section \ref{global_design}). With the received global prototypes, we define a regularization term (a part of the loss function), $\mathcal{L}_n^\textrm{RE}$, based on mean squared error (MSE) as follows:
\begin{subequations}\label{Opt_9}
	\begin{align}
		\mathcal{L}_n^\textrm{RE} = \frac{1}{|\mathcal{C}_n||\mathcal{P}_{n ,c}|}\sum_{c=1}^{|\mathcal{C}_n|} \sum_{i=1}^{|\mathcal{P}_{n ,c}|} (P_{n, c}^i-P_{c})^2. \tag{9}
	\end{align}
\end{subequations} 

\subsection{Local Objective}
\label{local_objective}
Regarding the local loss function, we employ a linear combination, which is formulated as follows:  
\begin{subequations}\label{Opt_10}
	\begin{align}
		\mathcal{L}_n^\textrm{loss} = \beta_1 \mathcal{L}_n^\textrm{CEO} + (1-\beta_1)\mathcal{L}_n^\textrm{KD}+\beta_2 \mathcal{L}_n^\textrm{CEA} + \notag \\ \beta_3 \mathcal{L}_n^\textrm{RE} + \beta_4 \mathcal{L}_n^\textrm{AF}, \tag{10}
	\end{align}
\end{subequations} 
where $\beta_1$, $\beta_2$, $\beta_3$ and $\beta_4$ are the hyperparameters. $\mathcal{L}_n^\textrm{CEO}$ is the CE loss based on the original data.

\subsection{Global Design}
\label{global_design}
On the server side, several processes will be conducted, i.e., global knowledge extraction (GKE), model aggregation, and multi-prototype aggregation. Note that GKE involves global geometric shapes (GGS) generation and global vectors (GVs) production. The details will be given as follows.

\par
\textbf{GGS Generation.} \quad
\label{global_geometric_shapes_generation} GGS per class can be obtained from the global covariance matrix (GCM) of each class $c$ (denoted as $\varrho^\textrm{c}$), where $\varrho^\textrm{c}$ can be calculated as follows \cite{ma2025geometric}: 
\begin{subequations}\label{Opt_11}
	\begin{align}
		\varrho^\textrm{c} = \frac{1}{M_c}\left(\sum_{n=1}^{|\mathcal{N}|}|\mathcal{D}_{n, c}|\varrho_n^\textrm{c} + \sum_{n=1}^{|\mathcal{N}|}|\mathcal{D}_{n, c}|(\mu_n^\textrm{c}-\mu^\textrm{c})(\mu_n^\textrm{c}-\mu^\textrm{c})^\top\right), \tag{11}
	\end{align}
\end{subequations} 
where $M_c = \sum_{n=1}^{|\mathcal{N}|}|\mathcal{D}_{n, c}|$ is the total number of all data samples belonging to class $c$. $\mathcal{N}$ denotes the set of satellites, while $|\mathcal{N}|$ is the overall number of satellites. As for $\mu^\textrm{c}$, it is the mean value of all the data, which can be computed by $\mu^\textrm{c} = \frac{1}{\sum_{n=1}^{|\mathcal{N}|}|\mathcal{D}_{n, c}|}\sum_{n=1}^{|\mathcal{N}|}\sum_{m=1}^{|\mathcal{D}_{n, c}|}x_n^\textrm{m,c} = \frac{1}{\sum_{n=1}^{|\mathcal{N}|}|\mathcal{D}_{n, c}|}\sum_{n=1}^{|\mathcal{N}|}|\mathcal{D}_{n, c}|\mu_n^c$. For the obtained $\varrho^\textrm{c}$, the eigenvalue decomposition process will be performed to yield eigenvalues (denoted as $\{\xi_1^\textrm{c}, \xi_2^\textrm{c}, \ldots, \xi_G^\textrm{c}\}$) and corresponding eigenvectors (denoted as $\{\upsilon_1^\textrm{c}, \upsilon_2^\textrm{c}, \ldots, \upsilon_G^\textrm{c}\}$), which can be combined to form the geometric shape \cite{ma2025geometric}. Here, $G$ is the number of the eigenvalues (or eigenvectors). Thus, the GGS of class $c$ for satellite $n$ is given as $GGS_c=\{\xi_1^\textrm{c}, \xi_2^\textrm{c}, \ldots, \xi_G^\textrm{c}, \upsilon_1^\textrm{c}, \upsilon_2^\textrm{c}, \ldots, \upsilon_G^\textrm{c}\}$. 

\begin{figure}[t!]
	\begin{algorithm}[H]	
		\renewcommand{\algorithmicrequire}{\textbf{Input:}}
		\renewcommand{\algorithmicensure}{\textbf{Output:}}
		\caption{Proposed GK-FedDKD Approach}
		\label{alg_1}
			\begin{algorithmic}[1]
				\REQUIRE $\mathcal{D}_n = \{(x_n^\textrm{1}, y_n^\textrm{1}), \ldots, (x_n^\textrm{m}, y_n^\textrm{m})\}$,
				\ENSURE Global Model $\omega$ \\
                \STATE Prepare TE for each local client via the first KD \\
				\hspace{-0.55cm} \textbf{Server-side Process:}
				\STATE Initialize the global model (i.e., $\omega$), a set of global prototypes (i.e., $\mathcal{P} = \{P_{1}, P_{2}, ..., P_{c}\}$) and $\Omega_c = \emptyset$
				\FOR{each round $t = 1, 2, ..., T$}
				\FOR{each participating satellite client}			
				\STATE $\varrho_n^\textrm{c}, \mu_n^\textrm{c}, \omega_n^t, \mathcal{P}_n \leftarrow \textrm{ClientUpdate}(\omega, \mathcal{P}, \Omega_c)$
				\ENDFOR
                \STATE Calculate the global covariance matrix $\varrho^\textrm{c}$ via \eqref{Opt_11} 
				\STATE Obtain global model $\omega$ 
                \STATE Perform the eigenvalue decomposition process on $\varrho^\textrm{c}$ to get $GGS_c$
                \STATE Obtain the global vectors of each class $\Omega_c$ via \eqref{Opt_12}
                \STATE Obtain global prototype set $\mathcal{P}$ via  \eqref{Opt_13}
				\ENDFOR
				\STATE \textbf{return} $\omega$
				\hfill \break
				\\
				\hspace{-0.63cm} \textbf{ClientUpdate}($\omega$, $\mathcal{P}, \Omega_c$):
				\STATE Update the local student model with $\omega$: $\omega_n \leftarrow \omega$
                \STATE Update the parameter of the classifier of the teacher network with the classifier of the student network
				\FOR{each epoch}
				\FOR{batch $(x_n, y_n) \in \mathcal{D}_n$}
				\STATE Obtain a multi-prototype of each category via K-Means
				\STATE Augment the local embedding with $\Omega_c$ via \eqref{Opt_5}
                \STATE Compute CE loss with original data: $\mathcal{L}_n^\textrm{CEO}$
				\STATE Compute KD loss $\mathcal{L}_n^\textrm{KD}$ via  \eqref{Opt_3}
				\STATE Compute CE loss with the augmented local embedding $\mathcal{L}_n^\textrm{CEA}$ via  \eqref{Opt_6}
				\STATE Compute  $\mathcal{L}_n^\textrm{AF}$ via  \eqref{Opt_7}
				\STATE Compute $\mathcal{L}_n^\textrm{RE}$ via \eqref{Opt_9}	
				\STATE Obtain local loss $\mathcal{L}_n^\textrm{loss}$ via \eqref{Opt_10}
				\STATE $\omega_n \leftarrow \omega_n - \hat{\delta}\bigtriangledown\mathcal{L}_n^\textrm{loss}$ \cite{mcmahan2017communication}
				\ENDFOR
				\ENDFOR	
				\STATE Get the updated local student model ($\omega_n^t$) and a set of multi-prototype ($\mathcal{P}_n = \{\{P_{n, c}^i\}_{i=1}^{|\mathcal{P}_{n, c}|}, c \in \mathcal{C}_n\}$) at round $t$
                \STATE Compute $\varrho_n^\textrm{c}$ and $\mu_n^\textrm{c}$
				\STATE \textbf{return} $\varrho_n^\textrm{c}$, $\mu_n^\textrm{c}$, $\omega_n^t$, $\mathcal{P}_n$
			\end{algorithmic}  
		\end{algorithm} 
        \vspace{-0.5cm}
	\end{figure}
\par
\textbf{GVs Production.} \quad As mentioned before, the obtained $GGS_c$ will be utilized to generate a unified global vector of class $c$, $\Omega_c$, for all clients to augment the local data. Particularly, based on $GGS_c$, we define $\Omega_c$ as follows:
\begin{subequations}\label{Opt_12}
	\begin{align}
		\Omega_c = \alpha\sum_{g=1}^{G}\xi_g^c\upsilon_g^c, \tag{12}
	\end{align}
\end{subequations} 
where $\alpha \sim N(0, 1)$, $N(0, 1)$ denotes normal distribution, and $E$ is the total number of eigenvectors/eigenvalues.


\par
\textbf{Model Aggregation.} \quad
On the server side, we perform a model aggregation process by using FedAvg \cite{mcmahan2017communication} to generate the global model, which is given by $\omega = \sum_{n=1}^{|\mathcal{N}|}\frac{|\mathcal{D}_n|}{D}\omega_n^t$. Here, $D = \sum_{n=1}^{|\mathcal{N}|}|\mathcal{D}_n|$ is the overall number of data samples owned by all participating satellites, while $\omega_n^t$ indicates the satellite $n$'s local model at round $t$. 

\textbf{Multi-prototype Aggregation.} \quad As aforementioned, the local prototypes will be disseminated to the server for multi-prototype aggregation, which is defined as follows:

\begin{subequations}\label{Opt_13}
	\setlength{\abovedisplayskip}{3pt}
	\setlength{\belowdisplayskip}{3pt}
	\begin{align}
		P_{c} = \frac{1}{|\mathcal{N}_c||\mathcal{P}_{n, c}|}\sum_{n=1}^{\left |\mathcal{N}_c\right |}\sum_{i=1}^{|\mathcal{P}_{n, c}|} \frac{\left| \mathcal{D}_{n, c} \right|}{\left| \mathcal{D}_c \right|}P_{n, c}^{i}.  \tag{13} 
	\end{align}	
\end{subequations}
where $|\mathcal{N}_c|$ denotes the number of satellites that own data belonging to class $c$. $\mathcal{D}_c=\bigcup_{n=1}^{|\mathcal{N}_c|}\mathcal{D}_{n, c}$ represents the set of data that belongs to class $c$ in the whole system. $|\mathcal{D}_c|$ is the size of the data corresponding to class $c$.

\subsection{Summary of Algorithm}
\par
In Algorithm \ref{alg_1}, we give the summary of the proposed GK-FedDKD approach. Concretely, in line $1$, the teacher encoder of each local client is generated by using the first KD (see Section \ref{Teacher_Encoder_Generation_KD_No_Labels}). From lines $2-13$, the server-side process is provided. Specifically, in line $2$, the global model, the global prototypes, and the global vectors of each class $c$ (i.e., $\Omega_c$) will be initialized. Then, for each round (line $3$), each participating client (line $4$) will perform a local update (line $5$). The global covariance matrix will be computed in line $7$, while the global model will be obtained in line $8$. In line $9$, we perform the eigenvalue decomposition process on the GCM per class $\varrho^\textrm{c}$ to get $GGS_c$. In line $10$, the global vectors of each class will be executed. In line $11$, the global prototype set will be obtained. The process regarding the local client update is provided in lines $14-31$. Particularly, we update the local student model with the global model in line $14$. Since we consider the shared classifier for both the student network and the teacher network, we update the parameter of the classifier of the teacher network with the parameter of the student network's classifier in line $15$. In each epoch (line $16$), for each batch (line $17$), we use K-Means to get the multi-prototype per category in line $18$. In line $19$, we augment the local data embedding with the global vectors. In lines $20-24$, we calculate $\mathcal{L}_n^\textrm{CEO}, \mathcal{L}_n^\textrm{KD}, \mathcal{L}_n^\textrm{CEA}, \mathcal{L}_n^\textrm{AF}$, and $\mathcal{L}_n^\textrm{RE}$ to form the local loss $\mathcal{L}_n^\textrm{loss}$ (line $25$). In line $26$, we update the local model by using $\omega_n \leftarrow \omega_n - \hat{\delta}\bigtriangledown\mathcal{L}_n^\textrm{loss}$ \cite{mcmahan2017communication}, where $\hat{\delta}$ indicates the learning rate. In line $29$, the updated local student model (i.e., $\omega_n^t$) and the multi-prototype set (i.e., $\mathcal{P}_n$) at round $t$ will be generated. In line $30$, we will obtain $\varrho_n^\textrm{c}$ and $\mu_n^\textrm{c}$. Finally, in line $31$, the obtained $\varrho_n^\textrm{c}$, $\mu_n^\textrm{c}$, $\omega_n^t$, and $\mathcal{P}_n$ will be sent to the server.

\subsection{Time Complexity}
The time complexity of the proposed GK-FedDKD method is provided according to the following primary aspects for simplicity: 1) model aggregation on the server, 2) multi-prototype aggregation on the server, 3) GCM calculation on the server, 4) the eigenvalue decomposition process on the server, 5) TE generation process on each local client (pre-training stage), 6) multi-prototype generation on each local satellite client, 7) LCM calculation on each local client, and 8) the underling network structure (i.e., backbone network) for performing the non-IID data classification task. For the model aggregation process, FedAvg is used. Considering there are total $|\mathcal{N}|$ satellites join the training, as per \cite{li2024fedgta}, the time complexity for conducting FedAvg on the server is bounded as $O(|\mathcal{N}|)$. As for the process of multi-prototype aggregation on the server, prior to analyzing the time complexity of that process, it is necessary to introduce the single prototype policy, where only a single prototype for a class in each local client. Let $\mathcal{C} = \mathcal{C}_1 \cup \mathcal{C}_2 \cup \ldots \cup \mathcal{C}_n$ denote the set of class in the global view, namely, all the categories across all the clients, and let $|\mathcal{C}|$ represent the number of classes, we can define the time complexity of the prototype aggregation via single prototype policy to be not greater than $O(|\mathcal{C}||\mathcal{N}|)$ \cite{zou2024cyber}. Accordingly, the time complexity for the multi-prototype aggregation of this work can be deduced to be no more than $O(|\mathcal{C}|\sum_{n=1}^{|\mathcal{N}|}|\mathcal{P}_{n, c}|)$. Regarding the GCM per class $c$ (i.e., $\varrho^\textrm{c}$) calculation on the server, each $\varrho^\textrm{c}$ is computed by using eq. \eqref{Opt_11}, the time complexity of the first part of eq. \eqref{Opt_11} (i.e., the part before ``+") can be given as $O(|\mathcal{N}|)$. The time complexity of the second part of eq. \eqref{Opt_11} (i.e., the part after ``+") can be defined as $O(|\mathcal{N}|s^2)$ \cite{magdon2010approximating, cui2009privacy}, where $s$ is the dimension. Hence, for all the categories, the overall time complexity of the server-side GCMs calculation can be gained by $O\Big(|\mathcal{C}| \big(|\mathcal{N}| + |\mathcal{N}|s^2\big)\Big)$. With regard to the server-side eigenvalue decomposition process, the time complexity is $O(s^3)$ to decompose each $\varrho^\textrm{c}$ \cite{li2014large} \cite{tufts2003simple} \cite{li2010making}. Since we consider total $|\mathcal{C}|$ categories, the time complexity for the eigenvalue decomposition process can be calculated as $O(|\mathcal{C}|s^3)$. Consequently, the server's time complexity can be given to be no greater than $O\Big(|\mathcal{N}| + |\mathcal{C}|\sum_{n=1}^{|\mathcal{N}|}|\mathcal{P}_{n, c}| + |\mathcal{C}| \big(|\mathcal{N}| + |\mathcal{N}|s^2\big) + |\mathcal{C}|s^3\Big)$. In the case of the TE generation on each local client, it relies on the underlying network structure being used. For convenience, a convolutional neural network (CNN) is used as an example to illustrate the time complexity of the TE and the SEs. For TE and each SE, CNN's time complexity can be given as $O(\sum_{j} \Phi_{j-1} \cdot \varsigma_j^2 \cdot \Phi_j \cdot b_j^2)$ \cite{he2015convolutional}. Here, $j$ is used to indicate the index of the convolutional layer. $\Phi_{j-1}$ and $\Phi_j$ are utilized to represent the input channels' number of the $j^\textrm{th}$ layer, and the filters' number in the $j^\textrm{th}$ layer. $\varsigma_j$ represents the length of the filter, while $b_j$ indicates the output feature map's length. With respect to the multi-prototype generation on each local satellite client, as aforementioned, K-Means is employed. According to \cite{fahim2006efficient, pakhira2014linear}, the time complexity of generating local prototypes from the data belonging to class $c$ on each local client by K-Means can be defined as $O(|\mathcal{D}_{n, c}||\mathcal{P}_{n, c}|\widehat{T})$. Here, $\widehat{T}$ represents the iteration number for completing the process of clustering. Hence, the time complexity for using K-Means to generate multiple prototypes for all the classes per client $n$ is $O(\sum_{c=1}^{|\mathcal{C}_n|}|\mathcal{D}_{n, c}||\mathcal{P}_{n, c}|\widehat{T})$. In regard to the calculation of the LCM of all classes on each local client, based on \cite{magdon2010approximating, cui2009privacy}, the time complexity can be obtained by $O(|\mathcal{C}_n||\mathcal{D}_{n, c}|s^2)$. Concerning the time complexity of the backbone network for classifying the non-IID data, it leans on the utilized neural network. For simplicity, as an example of the backbone network for both the TN and the SN, the combination of a CNN and a fully connected network (FCN) is considered. As aforementioned, the CNN's time complexity is defined as $O(\sum_{j} \Phi_{j-1} \cdot \varsigma_j^2 \cdot \Phi_j \cdot b_j^2)$. For the time complexity of the FCN, it can be obtained by $O(\sum_z\hat{\Phi}_{z-1}\hat{\Phi}_z)$, where $\hat{\Phi}_{z-1}$ and $\hat{\Phi}_z$ are utilized to denote the number of the neural units of FCN's $z-1^\textrm{th}$ and $z^\textrm{th}$ layer, respectively. As a result, in addition to the pre-training stage for generating TE, inspired by \cite{zou2024cyber}, the TN's time complexity for each client is not more than $O\Big(E\big(\sum_{j} \Phi_{j-1} \cdot \varsigma_j^2 \cdot \Phi_j \cdot b_j^2 + \sum_z\hat{\Phi}_{z-1}\hat{\Phi}_z \big) + |\mathcal{C}_n||\mathcal{D}_{n, c}|s^2 + |\mathcal{N}| + |\mathcal{C}|\sum_{n=1}^{|\mathcal{N}|}|\mathcal{P}_{n, c}| + |\mathcal{C}| \big(|\mathcal{N}| + |\mathcal{N}|s^2\big) + |\mathcal{C}|s^3 \Big)$, where $E$ is leveraged to denote the local iteration. As for the SN's time complexity, it is not higher than $O\Big(E(\sum_{j} \Phi_{j-1} \cdot \varsigma_j^2 \cdot \Phi_j \cdot b_j^2 + \sum_z\hat{\Phi}_{z-1}\hat{\Phi}_z + \sum_{c=1}^{|\mathcal{C}_n|}|\mathcal{D}_{n, c}||\mathcal{P}_{n, c}|\widehat{T}) + |\mathcal{C}_n||\mathcal{D}_{n, c}|s^2 + |\mathcal{N}| + |\mathcal{C}|\sum_{n=1}^{|\mathcal{N}|}|\mathcal{P}_{n, c}| +  |\mathcal{C}| \big(|\mathcal{N}| + |\mathcal{N}|s^2\big) + |\mathcal{C}|s^3 \Big)$.




\subsection{Convergence Analysis}
In this section, we will conduct a convergence analysis of the proposed GK-FedDKD approach. Prior to analyzing the convergence, several symbols are illustrated beforehand. As defined before, we use $E$ to define the total local iterations, and use $t$ to denote the communication round. In addition to this, we introduce $e \in \{\psi, 1, 2, \ldots, E\}$ to indicate the local iteration, where $\psi \in (0, 1)$ is utilized to denote the time step between the server completes the multi-prototype aggregation process and the start of the first local iteration. Accordingly, $tE+e$ can be leveraged to indicate the $e^\textrm{th}$ local iteration when the communication round is $t+1$. $tE+\psi$ is the time step before starting the first local iteration, after the server performs the multi-prototype aggregation. $tE$ denotes the local iteration before the server performs multi-prototype aggregation. Next, the details regarding the convergence analysis will be elaborated.
\par
Referring to \cite{tan2022fedproto, li2025federated, wu2022node, chang2025gpafed}, it is necessary to introduce Assumption $1$ and Assumption $2$.
\par
\textbf{Assumption $1$ (Convex, Lipschitz Smooth):} \quad  The local loss function (i.e., $\mathcal{L}_n^\textrm{loss}$) is convex and $L$-Lipschitz smooth for each client $n$. Hence, we can get the following inequality: 
\begin{subequations}\label{Opt_14}
	\setlength{\abovedisplayskip}{3pt}
	\setlength{\belowdisplayskip}{3pt}
	\begin{align}
         \lVert \bigtriangledown \mathcal{L}_{n,t_1}^\textrm{loss} - \bigtriangledown \mathcal{L}_{n,t_2}^\textrm{loss} \rVert  &\leq L \lVert  \omega_{n,t_1} - \omega_{n,t_2} \rVert, 
        \forall t_1, t_2 > 0, n \in \mathcal{N}.
        \tag{14} 
	\end{align}	
\end{subequations}
Here, $\lVert \cdot \rVert$ represents the $\ell_2$-norm \cite{chang2025gpafed}. As per \cite{tan2022fedproto}, for $\forall t_1, t_2>0$, eq. \eqref{Opt_14} can imply
\begin{subequations}\label{Opt_15}
	\setlength{\abovedisplayskip}{3pt}
	\setlength{\belowdisplayskip}{3pt}
	\begin{align}
         \mathcal{L}_{n,t_1}^\textrm{loss} - \mathcal{L}_{n,t_2}^\textrm{loss} \leq   
         \langle \bigtriangledown \mathcal{L}_{n,t_2}^\textrm{loss}, (\omega_{n,t_1} - \omega_{n,t_2})\rangle + \notag \\ \frac{L}{2}\lVert  \omega_{n,t_1} - \omega_{n,t_2} \rVert^2.
        \tag{15} 
	\end{align}	
\end{subequations}
In eq. \eqref{Opt_15}, $\langle \cdot,\cdot \rangle$ represents the inner product \cite{dinh2020federated}. If $L>0$, then $\mathcal{L}_n^\textrm{loss}$ is $L$-strongly convex for $\forall t_1, t_2$ \cite{chai2021fedat}.
\par
\textbf{Assumption $2$ (Unbiased Gradient \& Bounded Variance):} \quad For each client $n$, the stochastic gradient (denoted as $g_{n,t}=\bigtriangledown \mathcal{L}_n^\textrm{loss}(\omega_{n,t}, \Psi)$) is unbiased \cite{li2025federated}, where $\Psi$ indicates a mini-batch of data gained by random sampling from $\mathcal{D}_n$ \cite{wang2021novel}. Same as \cite{tan2022fedproto}, we assume the expectation of $g_{n,t}$ as $\mathbb{E}_{\Psi \sim \mathcal{D}_n}[g_{n,t}]=\bigtriangledown \mathcal{L}_n^\textrm{loss}(\omega_{n,t})=\bigtriangledown \mathcal{L}_{n,t}^\textrm{loss}$, which has a bounded variance (i.e., $\mathbb{E}[\lVert g_{n,t}-\bigtriangledown \mathcal{L}_n^\textrm{loss}(\omega_{n,t})\rVert^2] \leq \sigma^2$, where $\sigma^2 \geq 0$).
\par
To analyze the convergence of this work, the following Assumption $3$ needs to be given.
\par
\textbf{Assumption $3$ (Bounded $\ell_2$-norm of Each Local Prototype and Each Global Prototype):} \quad For the $\ell_2$-norm of each local prototype  $P_{n, c}^i$ and each global prototype $P_{c}$, it is bounded by $\widehat{R}$, namely, we have
\begin{subequations}\label{Opt_16}
	\setlength{\abovedisplayskip}{3pt}
	\setlength{\belowdisplayskip}{3pt}
	\begin{align}
         \lVert P_{n, c}^i \rVert, \lVert P_{c} \rVert \leq \widehat{R}, n \in \mathcal{N}, c \in \mathcal{C}.
        \tag{16} 
	\end{align}	
\end{subequations}

\par
\par
According to Assumption $1$ and Assumption $2$, the following Lemma can be given based on \cite{tan2022fedproto}.
\begin{lemma} 
\label{lemma_1}
For an arbitrary client $n$, from the start of the communication round $t+1$ to the final local iteration, before the server conducts the multi-prototype aggregation process, the local loss function $\mathcal{L}_n^\textrm{loss}$ is bounded by:

\begin{subequations}\label{Opt_17}
	\setlength{\abovedisplayskip}{3pt}
	\setlength{\belowdisplayskip}{3pt}
	\begin{align}
         \mathbb{E}[\mathcal{L}_{n, (t+1)E}^\textrm{loss}] & \leq   \mathcal{L}_{n, tE+\psi}^\textrm{loss} - \notag \\  &  (\hat{\delta}-\frac{L\hat{\delta}^2}{2})\sum_{e=\psi}^{E-1}\lVert\bigtriangledown \mathcal{L}_{n, tE+e}^\textrm{loss}\rVert^2+ \frac{LE\hat{\delta}^2}{2}\sigma^2.
        \tag{17} 
	\end{align}	
\end{subequations}

\begin{IEEEproof}
Consider $\omega_n = \omega_n - \hat{\delta}\bigtriangledown g_{n,t}$ and eq. \eqref{Opt_15} in Assumption $1$, we can get

\begin{subequations}\label{Opt_18}
	\setlength{\abovedisplayskip}{3pt}
	\setlength{\belowdisplayskip}{3pt}
	\begin{align}
          \mathcal{L}_{n, tE+1}^\textrm{loss} \leq \mathcal{L}_{n, tE+\psi}^\textrm{loss} + \langle \bigtriangledown \mathcal{L}_{n,tE+\psi}^\textrm{loss}, (\omega_{n,tE+1} - \omega_{n,tE+\psi})\rangle + \notag \\ \frac{L}{2}\lVert  \omega_{n,tE+1} - \omega_{n,tE+\psi} \rVert^2 \notag \\  
          = \mathcal{L}_{n, tE+\psi}^\textrm{loss} -\hat{\delta}\langle \bigtriangledown \mathcal{L}_{n,tE+\psi}^\textrm{loss}, g_{n, tE+\psi} \rangle + \frac{L}{2} \lVert \hat{\delta}g_{n,tE+\psi} \rVert^2.
        \tag{18} 
	\end{align}	
\end{subequations}
Taking the expectation on both sides of eq. \eqref{Opt_18}, as per \cite{tan2022fedproto}, we can yield
\begin{subequations}\label{Opt_19}
	\setlength{\abovedisplayskip}{3pt}
	\setlength{\belowdisplayskip}{3pt}
	\begin{align}     
        & \mathbb{E}[\mathcal{L}_{n,tE+1}] \leq\mathcal{L}_{n,tE+\psi}^\textrm{loss}-\hat{\delta}\mathbb{E}[\langle \bigtriangledown \mathcal{L}_{n,tE+\psi}^\textrm{loss}, g_{n, tE+\psi} \rangle] + \notag \\  
        & \qquad \qquad \qquad \qquad \qquad \qquad \qquad \qquad\frac{L\hat{\delta}^2}{2} \mathbb{E}[\lVert g_{n,tE+\psi} \rVert^2]  \notag \\
        & = \mathcal{L}_{n,tE+\psi}^\textrm{loss} - \hat{\delta}\lVert \bigtriangledown \mathcal{L}_{n,tE+\psi}^\textrm{loss} \rVert^2 + \frac{L\hat{\delta}^2}{2} \mathbb{E}[\lVert g_{n,tE+\psi} \rVert^2]  \notag \\
        & \leq \mathcal{L}_{n,tE+\psi}^\textrm{loss} - \hat{\delta}\lVert \bigtriangledown \mathcal{L}_{n,tE+\psi}^\textrm{loss} \rVert^2 + \notag \\ 
        &\qquad \qquad \qquad \frac{L\hat{\delta}^2}{2}\big(\lVert \bigtriangledown \mathcal{L}_{n,tE+\psi}^\textrm{loss} \rVert^2+ Var(g_{n,tE+\psi})\big) \notag \\ 
        & = \mathcal{L}_{n,tE+\psi}^\textrm{loss} \!- \!(\hat{\delta} \!-\!\frac{L\hat{\delta}^2}{2})\lVert \bigtriangledown \mathcal{L}_{n,tE+\psi}^\textrm{loss} \rVert^2 \!\!+\!\! \frac{L\hat{\delta}^2}{2}Var(g_{n,tE+\psi}) \notag \\
        & \leq \mathcal{L}_{n,tE+\psi}^\textrm{loss} \!- \!(\hat{\delta} \!-\!\frac{L\hat{\delta}^2}{2})\lVert \bigtriangledown \mathcal{L}_{n,tE+\psi}^\textrm{loss} \rVert^2 + \frac{L\hat{\delta}^2}{2}\sigma^2.     
        \tag{19} 
	\end{align}	
\end{subequations}
Telescoping all the E local iterations, we can get eq. \eqref{Opt_17}, thereby Lemma \ref{lemma_1} is proved.
\end{IEEEproof}
\end{lemma}

\begin{figure*}[t!]
\centering
\begin{subfigure}{0.32\textwidth} 
\includegraphics[width=1.\linewidth]{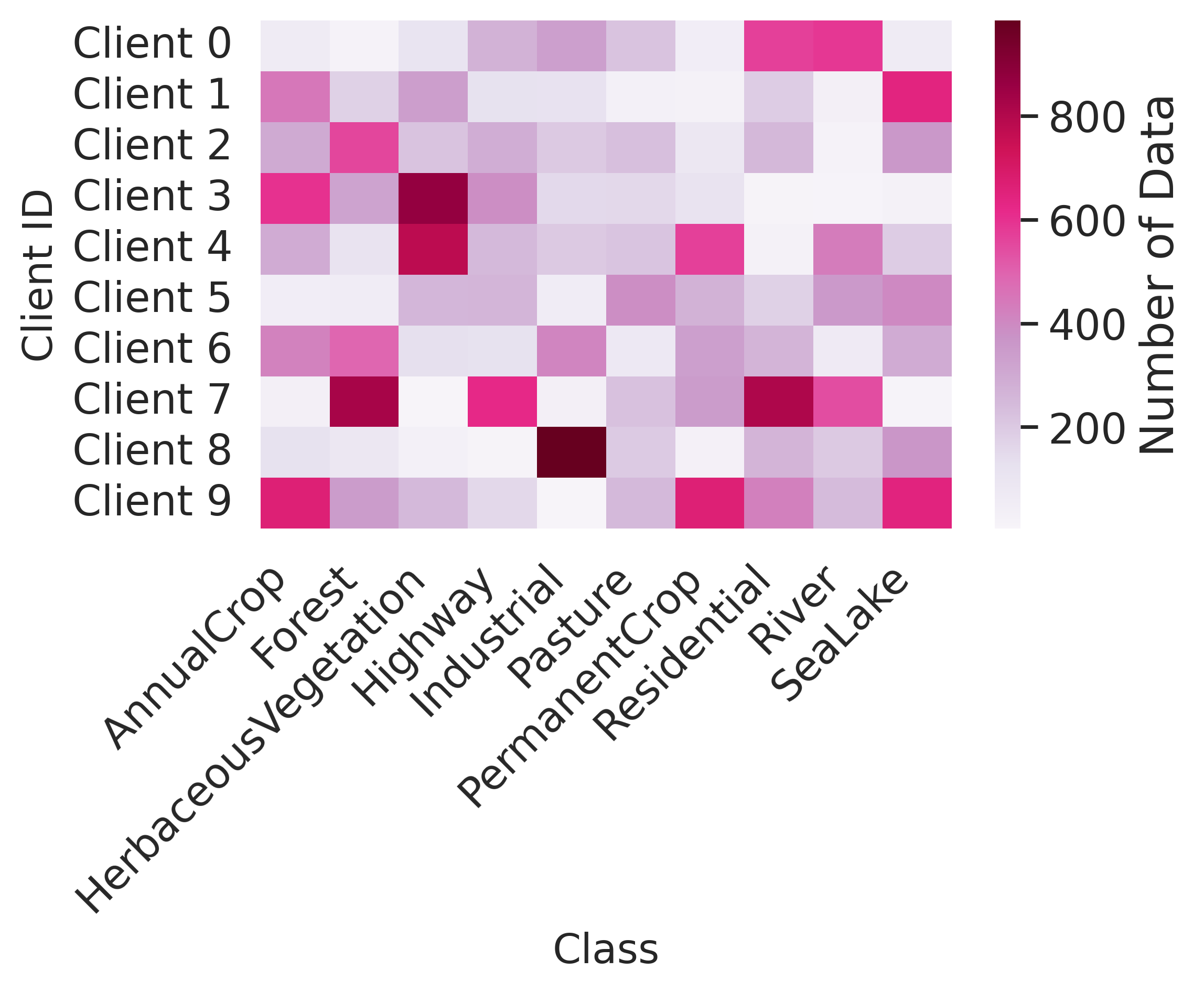}
\captionsetup{font=small} 
\caption{EuroSAT}
\label{EuroSAT_distribution}
\end{subfigure} \hfill
\begin{subfigure}{0.32\textwidth} 
\includegraphics[width=1.\linewidth]{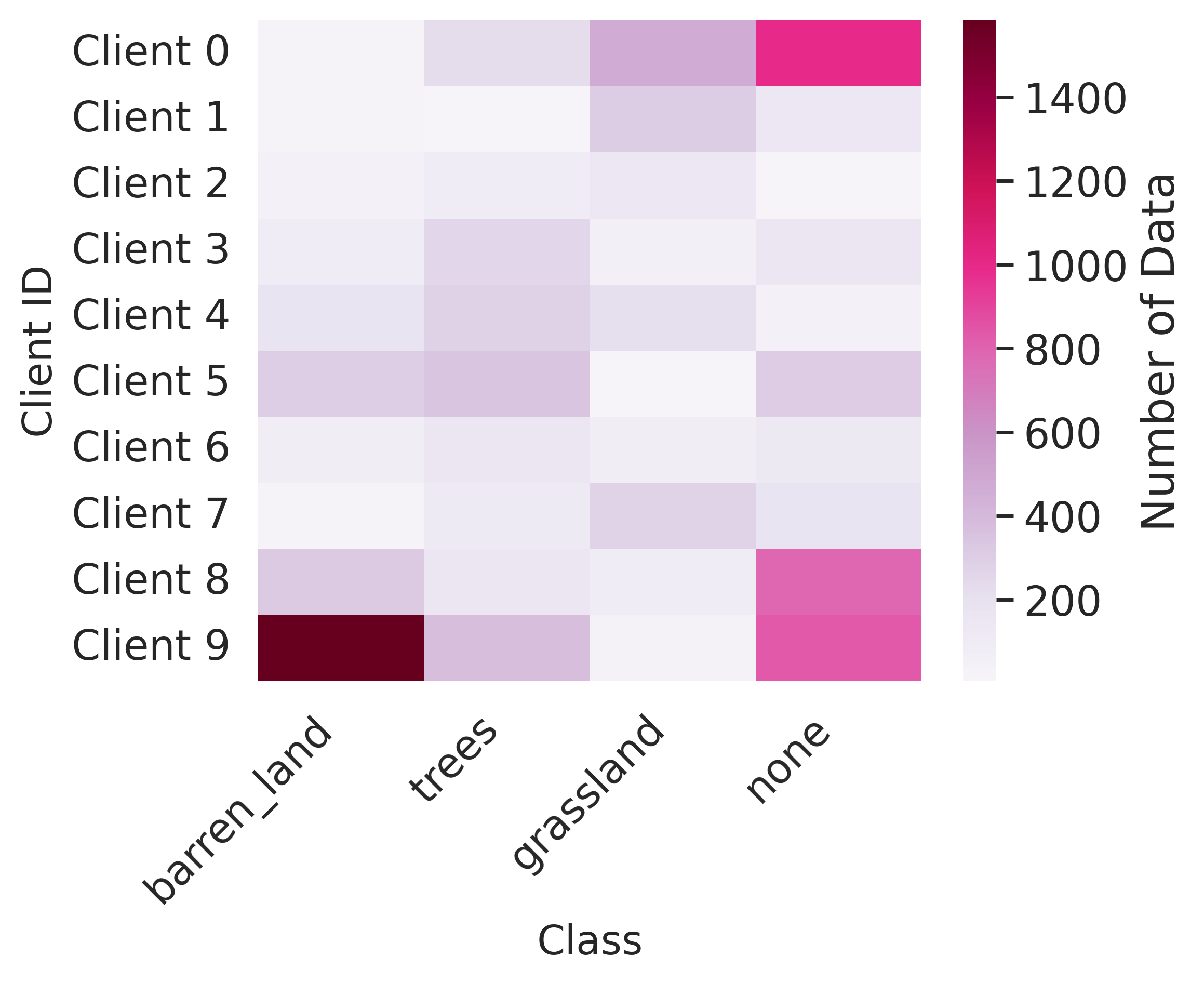}
\captionsetup{font=small} 
\caption{SAT4 (Train)}
\label{SAT4_train_distribution}
\end{subfigure} \hfill
\begin{subfigure}{0.32\textwidth} 
\includegraphics[width=1.\linewidth]{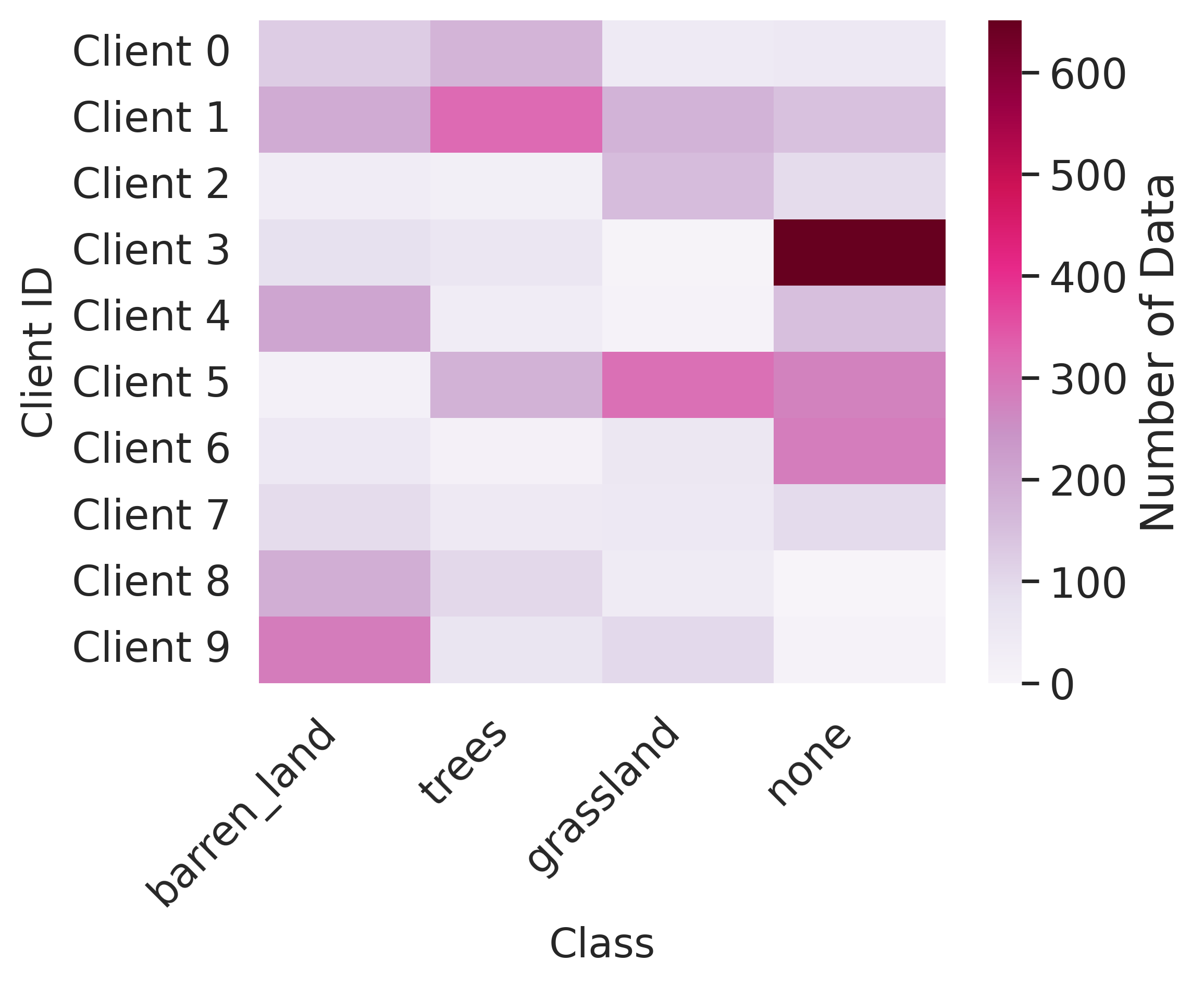}
\captionsetup{font=small} 
\caption{SAT4 (Test)}
\label{SAT4_test_distribution}
\end{subfigure} \\
\begin{subfigure}{0.32\textwidth} 
\includegraphics[width=1.\linewidth]{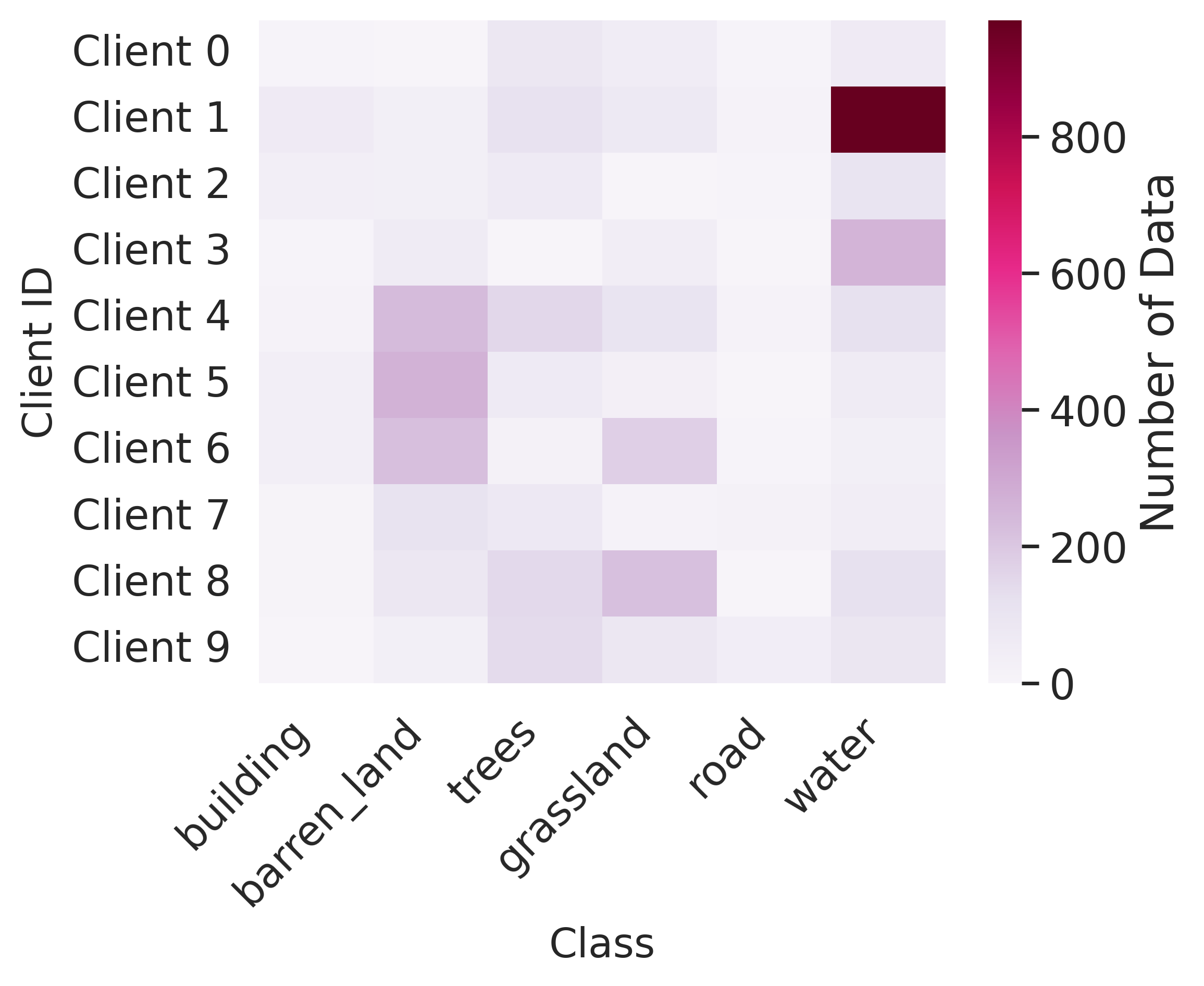}
\captionsetup{font=small} 
\caption{SAT6 (Train)}
\label{SAT6_train_distribution}
\end{subfigure} \hfill
\begin{subfigure}{0.32\textwidth} 
\includegraphics[width=1.\linewidth]{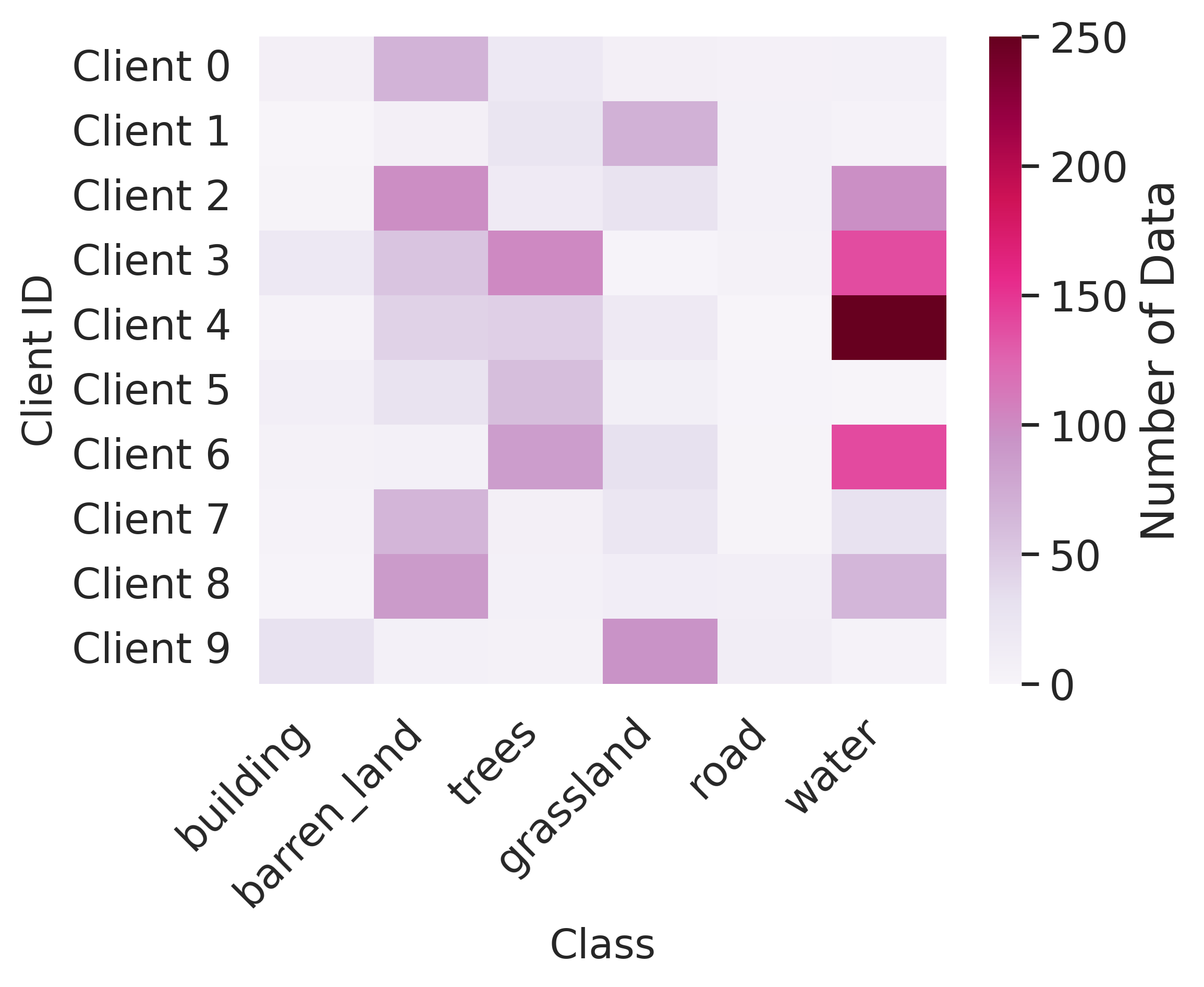}
\captionsetup{font=small} 
\caption{SAT6 (Test)}
\label{SAT6_test_distribution}
\end{subfigure} \hfill
\begin{subfigure}{0.32\textwidth} 
\includegraphics[width=1.\linewidth]{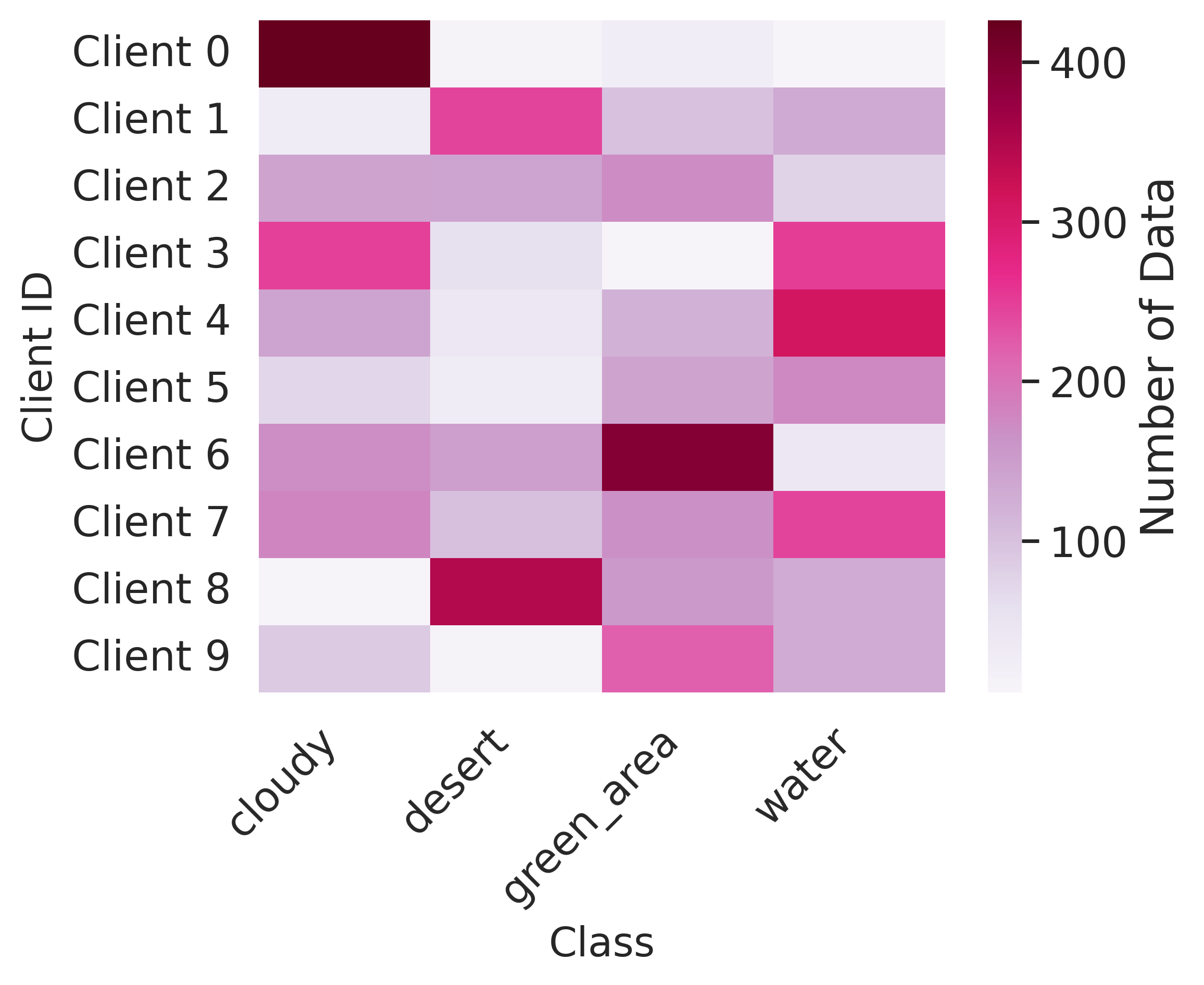}
\captionsetup{font=small} 
\caption{SIC}
\label{SIC_distribution}
\end{subfigure} 
\captionsetup{font=small} 
\caption {Per-class data sample distribution across ten clients.}
\label{data_statistics}
\end{figure*}

\par
Let Assumption 3 hold; we can get Lemma \ref{Opt_2} as follows:
\begin{lemma} 
\label{lemma_2}
After the multi-prototype aggregation process, for an arbitrary client $n$, the upper bound of the expectation of the local loss function is:
\begin{subequations}\label{Opt_20}
	\setlength{\abovedisplayskip}{3pt}
	\setlength{\belowdisplayskip}{3pt}
	\begin{align}
        \mathbb{E}[\mathcal{L}_{n,(t+1)E+\psi}^\textrm{loss}] \leq \mathcal{L}_{n,(t+1)E}^\textrm{loss} + 6\beta_3\widehat{R}^2.
        \tag{20} 
	\end{align}	
\end{subequations}
\begin{IEEEproof}
For $\mathcal{L}_{n,(t+1)E+\psi}^\textrm{loss}$, referring to \cite{tan2022fedproto} and eq. \eqref{Opt_10}, we have
\begin{subequations}\label{Opt_21}
	\setlength{\abovedisplayskip}{3pt}
	\setlength{\belowdisplayskip}{3pt}
	\begin{align}
          \mathcal{L}_{n,(t+1)E+\psi}^\textrm{loss}& = \mathcal{L}_{n,(t+1)E}^\textrm{loss} + \mathcal{L}_{n,(t+1)E+\psi}^\textrm{loss} - \mathcal{L}_{n,(t+1)E}^\textrm{loss} \notag \\
          & = \mathcal{L}_{n,(t+1)E}^\textrm{loss} + \beta_3\mathcal{L}_{n,(t+1)E+\psi}^\textrm{RE} - \beta_3\mathcal{L}_{n,(t+1)E}^\textrm{RE}.
        \tag{21} 
	\end{align}	
\end{subequations}
Considering eq. \eqref{Opt_9}, we can get
\begin{subequations}\label{Opt_22}
	\setlength{\abovedisplayskip}{3pt}
	\setlength{\belowdisplayskip}{3pt}
	\begin{align}
        & \beta_3\mathcal{L}_{n,(t+1)E+\psi}^\textrm{RE} - \beta_3\mathcal{L}_{n,(t+1)E}^\textrm{RE} \notag \\
        & = \beta_3\frac{1}{|\mathcal{C}_n||\mathcal{P}_{n ,c}|}\sum_{c=1}^{|\mathcal{C}_n|} \sum_{i=1}^{|\mathcal{P}_{n ,c}|} (P_{n, c, (t+1)E}^i-P_{c, t+2})^2 - \notag \\
        & \qquad \qquad  \beta_3\frac{1}{|\mathcal{C}_n||\mathcal{P}_{n ,c}|}\sum_{c=1}^{|\mathcal{C}_n|} \sum_{i=1}^{|\mathcal{P}_{n ,c}|} (P_{n, c, (t+1)E}^i-P_{c, t+1})^2 \notag \\
        & = \frac{\beta_3}{|\mathcal{C}_n||\mathcal{P}_{n ,c}|}\sum_{c=1}^{|\mathcal{C}_n|} \sum_{i=1}^{|\mathcal{P}_{n ,c}|} \big[2P_{n, c, (t+1)E}^i(P_{c, t+1}-P_{c, t+2}) + \notag \\
        & \qquad \qquad \qquad \qquad \qquad \qquad (P_{c, t+2})^2 - (P_{c, t+1})^2 \big] \notag \\
        &  \leq \frac{\beta_3}{|\mathcal{C}_n||\mathcal{P}_{n ,c}|}\sum_{c=1}^{|\mathcal{C}_n|} \sum_{i=1}^{|\mathcal{P}_{n ,c}|} \big[2P_{n, c, (t+1)E}^i(P_{c, t+1}-P_{c, t+2}) + \notag \\
        & \qquad \qquad \qquad \qquad \qquad \qquad (P_{c, t+2})^2 + (P_{c, t+1})^2 \big] \notag \\
        & \stackrel{(a)}{\leq} \frac{\beta_3}{|\mathcal{C}_n||\mathcal{P}_{n ,c}|}\sum_{c=1}^{|\mathcal{C}_n|} \sum_{i=1}^{|\mathcal{P}_{n ,c}|} \big[2P_{n, c, (t+1)E}^i(P_{c, t+1}-P_{c, t+2}) + 2\widehat{R}^2 \big] \notag \\
        & = \frac{2\beta_3}{|\mathcal{C}_n||\mathcal{P}_{n ,c}|}\sum_{c=1}^{|\mathcal{C}_n|} \sum_{i=1}^{|\mathcal{P}_{n ,c}|} \big( \langle P_{n, c, (t+1)E}^i, P_{c, t+1}-P_{c, t+2} \rangle + \widehat{R}^2\big)
        \tag{22} 
	\end{align}	
\end{subequations}
Here, since we use $(t+1)E+\psi$ to denote the time step after multi-prototype aggregation on the server-side but before the beginning of the first iteration of the next round on each local client, we use $P_{c, t+2}$ to represent the newly updated global prototype of class $c$, and use $P_{c, t+1}$ to denote the global prototype belonging to class $c$ before updating. In eq. \eqref{Opt_22}, $(a)$ follows from $(P_{c, t+2})^2 = \langle P_{c, t+2}, P_{c, t+2} \rangle = \lVert P_{c, t+2} \rVert^2$, $(P_{c, t+1})^2 = \langle P_{c, t+1}, P_{c, t+1}\rangle = \lVert P_{c, t+1} \rVert^2$ and Assumption $3$. Next, taking the absolute value of both sides of eq. \eqref{Opt_22}, we can obtain

\begin{subequations}\label{Opt_23}
	\setlength{\abovedisplayskip}{3pt}
	\setlength{\belowdisplayskip}{3pt}
	\begin{align}
        & |\beta_3\mathcal{L}_{n,(t+1)E+\psi}^\textrm{RE} - \beta_3\mathcal{L}_{n,(t+1)E}^\textrm{RE}| \notag \\
        & \stackrel{(b)}{\leq} \frac{2\beta_3}{|\mathcal{C}_n||\mathcal{P}_{n ,c}|}\sum_{c=1}^{|\mathcal{C}_n|} \sum_{i=1}^{|\mathcal{P}_{n ,c}|} \big( \lVert P_{n, c, (t+1)E}^i \rVert \lVert P_{c, t+1}-P_{c, t+2}\rVert + \widehat{R}^2\big) \notag \\
        & \stackrel{(c)}{\leq} \frac{2\beta_3\widehat{R}}{|\mathcal{C}_n||\mathcal{P}_{n ,c}|}\sum_{c=1}^{|\mathcal{C}_n|} \sum_{i=1}^{|\mathcal{P}_{n ,c}|} \big(\lVert P_{c, t+1}-P_{c, t+2}\rVert + \widehat{R}\big) \notag \\
        & \stackrel{(d)}{\leq} \frac{2\beta_3\widehat{R}}{|\mathcal{C}_n||\mathcal{P}_{n ,c}|}\sum_{c=1}^{|\mathcal{C}_n|} \sum_{i=1}^{|\mathcal{P}_{n ,c}|} \big(\lVert P_{c, t+1}\rVert + \lVert P_{c, t+2}\rVert + \widehat{R}\big) \notag \\
        & \stackrel{(e)}{\leq} \frac{2\beta_3\widehat{R}}{|\mathcal{C}_n||\mathcal{P}_{n ,c}|}\sum_{c=1}^{|\mathcal{C}_n|} \sum_{i=1}^{|\mathcal{P}_{n ,c}|} \big(\widehat{R} + \widehat{R} + \widehat{R}\big) \notag \\
        & = 6\beta_3\widehat{R}^2,
        \tag{23} 
	\end{align}	
\end{subequations}
where $(b)$ follows the Cauchy–Schwarz Inequality (for two vectors $u$ and $v$ belonging to an inner product space, $\left| \langle u, v\rangle \right| \leq \lVert u \rVert \lVert v \rVert$) \cite{axler2024linear}. Through Assumption $3$, $(c)$ can be obtained. For any two vectors $u$ and $v$, the Triangle inequality (i.e., $\lVert u + v \rVert \leq \lVert u \rVert \lVert v \rVert$) holds. Hence, we can know $\lVert u + (-v) \rVert = \lVert u - v \rVert \leq \lVert u \rVert \lVert -v \lVert$. Since $\lVert -v \lVert = \lVert v \lVert$, we can further get $\lVert u - v \rVert \leq \lVert u \rVert \lVert v \rVert$. Accordingly, $(d)$ can be obtained. Based on Assumption $3$, (e) can be acquired. Substituting eq. \eqref{Opt_23} into eq. \eqref{Opt_21}, we can get the upper bound of the local loss function as follows:
\begin{subequations}\label{Opt_24}
	\setlength{\abovedisplayskip}{3pt}
	\setlength{\belowdisplayskip}{3pt}
	\begin{align}
     \mathcal{L}_{n,(t+1)E+\psi}^\textrm{loss} \leq \mathcal{L}_{n,(t+1)E}^\textrm{loss} + 6\beta_3\widehat{R}^2.
        \tag{24} 
	\end{align}	
\end{subequations}
Take the expectation form, we can get eq. \eqref{Opt_20} in Lemma \ref{lemma_2}. Hence, Lemma \ref{lemma_2} is proved.
\end{IEEEproof}
\end{lemma}

\par
Considering Assumption $1$ to Assumption $3$ simultaneously, the following theorem can be given.
\begin{theorem} 
\label{Theorem_1}
  Let Assumption $1$ to Assumption $3$ hold; after every round, for any arbitrary client $n$, we can get
  \begin{subequations}\label{Opt_25}
	\setlength{\abovedisplayskip}{3pt}
	\setlength{\belowdisplayskip}{3pt}
	\begin{align}
        & \mathbb{E}[\mathcal{L}_{n,(t+1)E+\psi}^\textrm{loss}] \leq \mathcal{L}_{n, tE+\psi}^\textrm{loss} -  (\hat{\delta}-\frac{L\hat{\delta}^2}{2})\sum_{e=\psi}^{E-1}\lVert\bigtriangledown \mathcal{L}_{n, tE+e}^\textrm{loss}\rVert^2 \notag \\  
        &  \qquad  \qquad \qquad \qquad \qquad \qquad \qquad  + \frac{LE\hat{\delta}^2}{2}\sigma^2 + 6\beta_3\widehat{R}^2.
        \tag{25} 
	\end{align}	
\end{subequations}
\begin{IEEEproof}
Inspired by \cite{tan2022fedproto}, we consider taking the expectation on both sides of Lemma \ref{lemma_1} and Lemma \ref{lemma_2}, and performing the summation operation thereafter. Then Theorem \ref{Theorem_1} can be easily acquired. Therefore, the convergence of the local loss function $\mathcal{L}_n^\textrm{loss}$ (i.e., eq. \eqref{Opt_10}) holds.
\end{IEEEproof}
\end{theorem}

\begin{figure*}[t!]
	\centering
	\begin{subfigure}{0.24\textwidth} 
		\includegraphics[width=1.\linewidth]{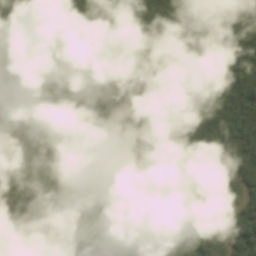}
		\captionsetup{font=small} 
		\caption{Original}
		\label{original_data}
	\end{subfigure} \hfill
	\begin{subfigure}{0.24\textwidth} 
		\includegraphics[width=1.\linewidth]{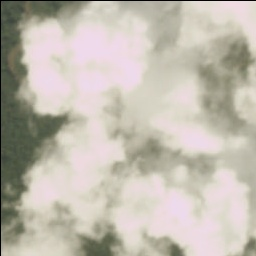}
		\captionsetup{font=small} 
		\caption{Rotation}
		\label{rotation_data}
	\end{subfigure} \hfill
	\begin{subfigure}{0.24\textwidth} 
		\includegraphics[width=1.\linewidth]{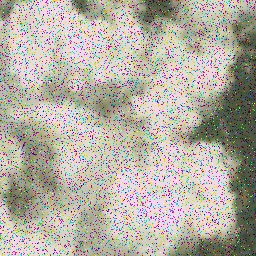}
		\captionsetup{font=small} 
		\caption{Gaussian Noise}
		\label{gaussian_noise_data}
	\end{subfigure} \hfil
	\begin{subfigure}{0.24\textwidth} 
		\includegraphics[width=1.\linewidth]{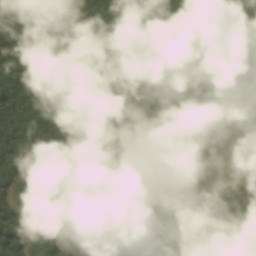}
		\captionsetup{font=small} 
		\caption{Flip}
		\label{flip_data}
	\end{subfigure} \\
	\begin{subfigure}{0.24\textwidth} 
		\includegraphics[width=1.\linewidth]{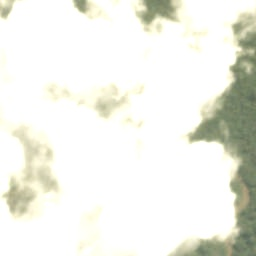}
		\captionsetup{font=small} 
		\caption{Brighter}
		\label{brighter_data}
	\end{subfigure} \hfill
	\begin{subfigure}{0.24\textwidth} 
		\includegraphics[width=1.\linewidth]{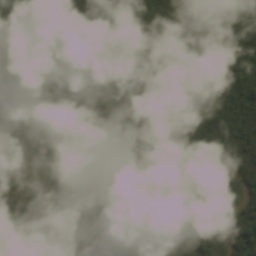}
		\captionsetup{font=small} 
		\caption{Darker}
		\label{darker_data}
	\end{subfigure} \hfill
	\begin{subfigure}{0.24\textwidth} 
		\includegraphics[width=1.\linewidth]{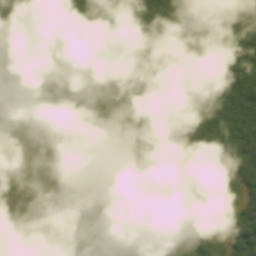}
		\captionsetup{font=small} 
		\caption{Saturation}
		\label{saturation_data}
	\end{subfigure} \hfil
	\begin{subfigure}{0.24\textwidth} 
		\includegraphics[width=1.\linewidth]{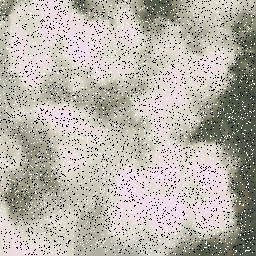}
		\captionsetup{font=small} 
		\caption{Salt and Pepper}
		\label{salt_and_pepper_data}
	\end{subfigure} 
	\captionsetup{font=small} 
	\caption {Data augmentation over the SIC dataset with rotation, Gaussian noise, flip, brighter, darker, saturation, and salt and pepper, respectively.}
	\label{data_augmentation_manner_SIC}
\end{figure*}

\begin{figure*}[h]
\centering
\begin{subfigure}{0.24\textwidth} 
\includegraphics[width=1.\linewidth]{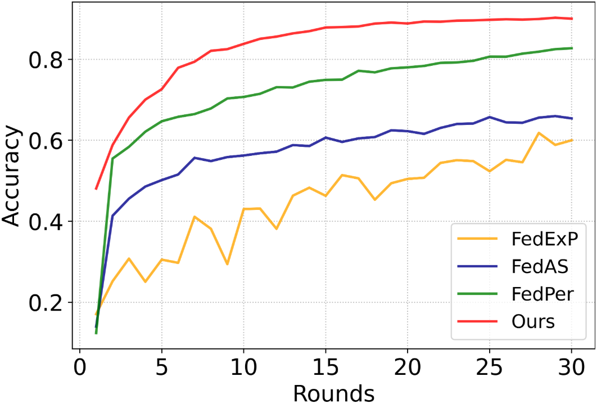}
\captionsetup{font=small} 
\caption{EuroSAT}
\label{EuroSAT_Convergence}
\end{subfigure} \hfill
\begin{subfigure}{0.24\textwidth} 
\includegraphics[width=1.\linewidth]{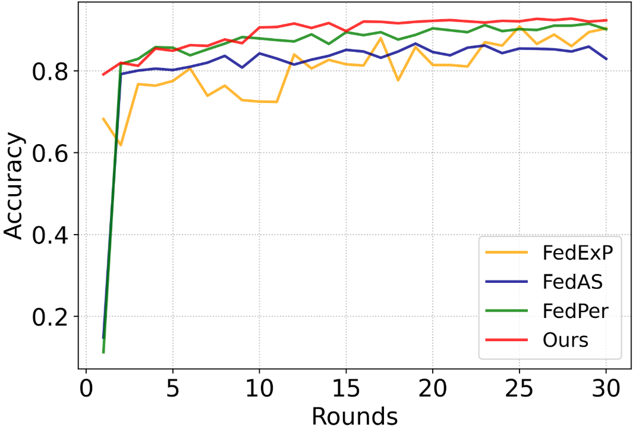}
\captionsetup{font=small} 
\caption{SIC}
\label{SIC_Convergence}
\end{subfigure} \hfill
\begin{subfigure}{0.24\textwidth} 
\includegraphics[width=1.\linewidth]{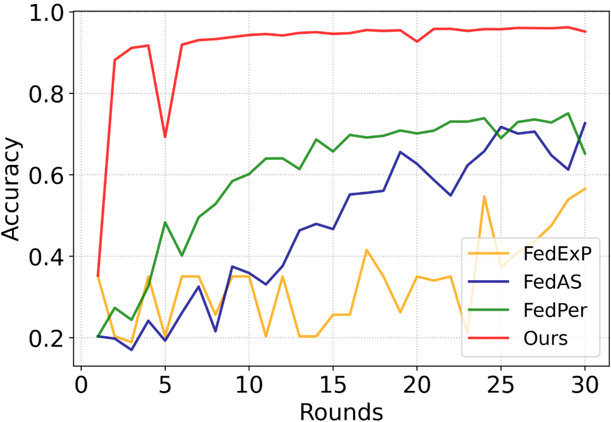}
\captionsetup{font=small} 
\caption{SAT4}
\label{SAT4_Convergence}
\end{subfigure} \hfil
\begin{subfigure}{0.24\textwidth} 
\includegraphics[width=1.\linewidth]{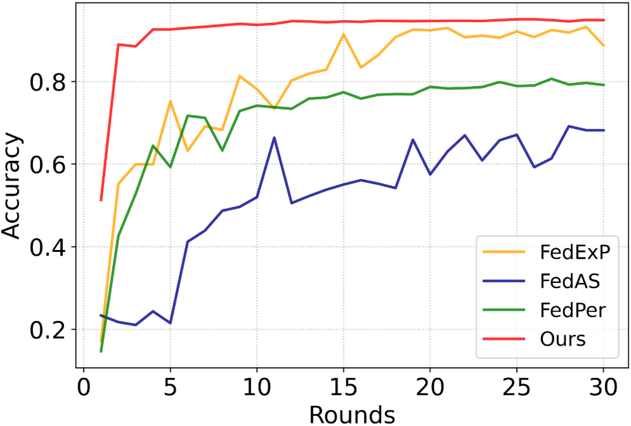}
\captionsetup{font=small} 
\caption{SAT6}
\label{SAT6_Convergence}
\end{subfigure} 
\captionsetup{font=small} 
\caption {Convergence comparison among FedExP, FedAS, FedPer and the proposed GK-FedDKD method.}
\label{convergence_comparison}
\end{figure*}

\begin{table*}[htpb]
\renewcommand{\thetable}{\Roman{table}}
\setlength{\extrarowheight}{1 pt}
\caption{Performance comparison among the proposed GK-FedDKD and other baselines (BB: Backbone; AA: Average Accuracy; MAS: Macro-F1 Score). The highest values are shown in bold, the second best is marked in blue, and the difference between the best and the second best is depicted in red.}
\begin{center}
\begin{tabular}{|p{0.07\textwidth}|p{0.12\textwidth}|>{\centering}p{0.055\textwidth}>{\centering}p{0.06\textwidth}>{\centering\arraybackslash}p{0.065\textwidth}||>{\centering}p{0.055\textwidth}>{\centering}p{0.057\textwidth}>{\centering\arraybackslash}p{0.062\textwidth}||>{\centering}p{0.055\textwidth}>{\centering}p{0.055\textwidth}>{\centering\arraybackslash}p{0.062\textwidth}|
}
	\hline	
	\centering \multirow{2}{*}{BB} & \centering \multirow{2}{*}{Method} & \multicolumn{3}{c||}{EuroSAT} & \multicolumn{3}{c||}{SAT4} & \multicolumn{3}{c|}{SAT6}\\ \cline{3-5} \cline{6-8} \cline{9-11}
	& & Accuracy &  AA & MFS & Accuracy & AA & MFS & Accuracy & AA & MFS \\
	\hline
	\centering \multirow{9}{*}{Swin-T}  & \centering FedExP \cite{jhunjhunwala2023fedexp} & $0.6001$  & $0.6046$  & $0.5877$ & $0.5656$ & $0.5896$ & $0.5206$ & $0.8875$ & $0.8652$ & $0.7301$ \\ \cline{2-11}
	& \centering MOON  \cite{li2021model} & $0.7179$ & $0.7187$ & $0.7016$ & $0.3506$ & $0.3194$ & $0.1298$ & $0.7335$ & $0.7084$ & $0.5310$ \\ \cline{2-11}
	&\centering FedAU \cite{wang2023lightweight} &  $0.6713$ & $0.6785$ & $0.6432$ & $0.7786$ & $0.7923$ & $0.7731$ & $\color{blue}{0.9240}$ & $\color{blue}{0.8997}$ & $\color{blue}{0.7913}$ \\ \cline{2-11} 
    & \centering FedProto \cite{tan2022fedproto} & $\color{blue}{0.8286}$ & $\color{blue}{0.8281}$ & $\color{blue}{0.8203}$ & $\color{blue}{0.8246}$ & $\color{blue}{0.8375}$ & $\color{blue}{0.8130}$ & $0.8740$ & $0.8678$ & $0.7495$ \\ \cline{2-11}
    &  \centering FedProx \cite{li2020federated} & $0.2156$ & $0.1348$ & $0.1294$ & $0.4992$ & $0.5257$ & $0.3955$ & $0.4075$ & $0.3247$ & $0.2067$ \\ \cline{2-11}
    &  \centering FedAS \cite{yang2024fedas} & $0.6541$ &  $0.6476$ & $0.6367$ & $0.7266$ & $0.7317$ & $0.7154$ & $0.7415$ & $0.7312$ & $0.5618$ \\ \cline{2-11}
    & \centering FedPer \cite{arivazhagan2019federated} & $0.8273$ & $0.8126$ & $\color{blue}{0.8203}$ & $0.6522$& $0.7108$ & $0.6553$ & $0.7915$ & $0.7696$ & $0.6420$\\ \cline{2-11}
    &  \centering \textbf{Ours} & $\textbf{0.9003}$ & $\textbf{0.9023}$ & $\textbf{0.8962}$ & $\textbf{0.9516}$ & $\textbf{0.9502}$ & $\textbf{0.9450}$ & $\textbf{0.9485}$ & $\textbf{0.9346}$ & $\textbf{0.8902}$ \\ \cline{2-11} 
    &  \centering $\color{red}{\bigtriangleup}$ & $\color{red}{+0.0717}$  & $\color{red}{+0.0742}$ & $\color{red}{+0.0759}$ & $\color{red}{+0.127}$ & $\color{red}{+0.1127}$ & $\color{red}{+0.132}$ & $\color{red}{+0.0245}$ & $\color{red}{+0.0349}$ & $\color{red}{+0.0989}$ \\ \hline \hline
    \centering \multirow{8}{*}{ResNet10}  & \centering FedExP \cite{jhunjhunwala2023fedexp} &  $0.1965$ & $0.2102$ & $0.0946$ & $0.4082$ & $0.3776$ & $0.2510$ & $0.1560$ & $0.1678$ & $0.1289$ \\ \cline{2-11}
	& \centering MOON  \cite{li2021model} & $0.5200$ & $0.5204$ & $0.4557$ & $\color{blue}{0.9250}$ & $\color{blue}{0.9227}$ & $\color{blue}{0.9179}$ & $0.8725$ & $0.8610$ & $0.8157$ \\ \cline{2-11}
	&\centering FedAU \cite{wang2023lightweight} & $0.7490$ & $0.7504$ & $0.7371$ & $0.8898$ & $0.8862$ & $0.8769$ & $0.9030$ & $0.8746$ & $\color{blue}{0.8195}$ \\ \cline{2-11} 
    & \centering FedProto \cite{tan2022fedproto} & $\color{blue}{0.8374}$ & $\color{blue}{0.8367}$ & $\color{blue}{0.8308}$ & $0.8852$ & $0.8943$ & $0.8752$ & $0.9000$ & $0.9027$ & $0.8147$ \\ \cline{2-11}
    &  \centering FedProx \cite{li2020federated}  & $-$ & $-$ & $-$ & $0.1892$ & $0.1907$ & $0.0795$  & $0.1470$ & $0.1912$ & $0.0428$ \\ \cline{2-11}
    &  \centering FedAS \cite{yang2024fedas}  & $0.6678$ & $0.6602$ & $0.6472$ & $0.8200$ & $0.8135$ & $0.7990$ & $\color{blue}{0.9290}$ & $\color{blue}{0.9130}$ & $0.7707$ \\ \cline{2-11}
    & \centering FedPer \cite{arivazhagan2019federated} & $0.7910$ & $0.7743$ & $0.7801$ & $0.7878$ & $0.7732$ & $0.7884$ & $0.8620$ & $0.8556$ & $0.7399$ \\ \cline{2-11}
    &  \centering \textbf{Ours} & $\textbf{0.8838}$ & $\textbf{0.8854}$ & $\textbf{0.8783}$ & $\textbf{0.9426}$ & $\textbf{0.9432}$ & $\textbf{0.9348}$ & $\textbf{0.9675}$ & $\textbf{0.9611}$ & $\textbf{0.9168}$ \\ \cline{2-11} 
    &  \centering $\color{red}{\bigtriangleup}$ & $\color{red}{+0.0464}$ & $\color{red}{+0.0487}$ & $\color{red}{+0.0475}$ & $\color{red}{+0.0176}$ & $\color{red}{+0.0205}$ & $\color{red}{+0.0169}$ & $\color{red}{+0.0385}$ & $\color{red}{+0.0481}$ & $\color{red}{+0.0973}$ \\ \hline 
\end{tabular}
\end{center}
\label{sec_4_tab1_accuracy_AA_MFS_Swin}
\end{table*}

\begin{table*}[t!]
\renewcommand{\thetable}{\Roman{table}}
\setlength{\extrarowheight}{1 pt}
\caption{Error rate and mean absolute error (MAE) achieved by the proposed method and other baselines. Bold numbers denote the best values, and blue numbers represent the second best. Notably, the \textit{\textbf{lower}} the value in this Table, the \textit{\textbf{better}} the result.}

\begin{center}
\begin{tabular}{p{0.07\textwidth}>{\centering}p{0.1\textwidth}>{\centering}p{0.12\textwidth}>{\centering\arraybackslash}p{0.1\textwidth}>{\centering}p{0.12\textwidth}>{\centering\arraybackslash}p{0.1\textwidth}>{\centering\arraybackslash}p{0.08\textwidth}>{\centering\arraybackslash}p{0.1\textwidth}}
	\hline \hline
	\multirow{2}{*}{Dataset} & \multirow{2}{*}{Metrics} &\multicolumn{6}{c}{Method}\\\cline{3-8}
	  &  & FedExP \cite{jhunjhunwala2023fedexp} & MOON \cite{li2021model} & FedProto \cite{tan2022fedproto} & FedAU \cite{wang2023lightweight}& Ours & \textbf{$\bigtriangleup$} \\ \hline
	\hline
	\centering \multirow{2}{*}{EuroSAT}  & Error Rate & $0.3999$ & $0.2821$ & $\color{blue}{0.1714}$ & $0.3287$ & $\textbf{0.0997}$ & $-0.0717$\\
    &  MAE & $1.7178$ & $1.1016$ & $\color{blue}{0.6726}$ & $1.3353$ & $\textbf{0.3710}$ & $-0.3016$\\ \hline 
    \centering \multirow{2}{*}{SIC} & Error Rate & $\color{blue}{0.0966}$ & $0.1572$ & $0.1452$ & $0.2406$ & $\textbf{0.0768}$ &  $-0.0198$ \\
    &  MAE & $\color{blue}{0.1044}$ & $0.1710$ & $0.1518$ & $0.2567$ & $\textbf{0.0840}$ & $-0.0204$ \\ \hline 
    \centering \multirow{2}{*}{SAT4} & Error Rate & $0.4344$ & $0.6494$ & $\color{blue}{0.1754}$ & $0.2214$ & $\textbf{0.0484}$ & $-0.127$ \\
    &  MAE & $0.747$ & $1.3662$ & $\color{blue}{0.2956}$  & $0.3954$ & $\textbf{0.0920}$ & $-0.2036$ \\ \hline 
    \centering \multirow{2}{*}{SAT6} & Error Rate & $0.1125$ & $0.2665$ & $0.1260$ & $\color{blue}{0.0760}$ & $\textbf{0.0515}$ & $-0.0245$\\
    &  MAE & $0.2355$ & $0.4835$ & $0.2500$ & $\color{blue}{0.1750}$ & $\textbf{0.1095}$ & $-0.0655$ \\ \hline \hline
\end{tabular}
\end{center}
\label{sec_4_tab2_Error_Comparison}
\end{table*}

\section{Experiments}
\label{sec_4_experiments}
In this section, the experimental analysis of the proposed method towards RSSI is conducted.
\par
\textbf{Dataset.} \quad We employ four satellite datasets for experiments: EuroSAT \cite{helber2019eurosat}, Satellite Image Classification (SIC) dataset \cite{satelliteClassficationDataset}, SAT4 \cite{basu2015deepsat}, and SAT6 \cite{basu2015deepsat}. For both EuroSAT ($27,000$ labeled images) and SIC, we randomly select $70\%$ as training data, while the remaining portion ($30\%$) is the test part. Regarding SAT4, for simplicity, we randomly select $10,000$ image patches for training and $5,000$ for testing. Regarding SAT6, for convenience, $5,000$ and $2,000$ image patches are randomly selected as training and test data, respectively.


\par
\textbf{Implementation Settings and Backbone.} \quad We give parameter settings as below: learning rate is $0.001$ unless otherwise specified, the batch size is $32$, the temperature $\tau$ is defined as $0.2$, round is set as $30$, local epochs are $10$, output embeddings of TE and SE are $32$. We also set the number of clients $10$ unless otherwise stated and adopted a random client selection mechanism during training. Besides, $\beta_1=0.9$, $\beta_2=0.1$, $\beta_3=\beta_4=0.01$, and the SGD optimizer are utilized in the FL setting. Note that to generate TE, we adopt the AdamW optimizer for each SE, and the local epochs are set to $40$. Each image is rescaled to $224 \times 224$ pixels, and a Dirichlet distribution with parameter $0.9$ \cite{zou2024cyber} is employed to prepare heterogeneous data (as shown in Fig. \ref{data_statistics}) unless otherwise indicated. In addition, traditional data augmentation methods (e.g., rotation, Gaussian noise, flip, brightness, contrast, saturation, and salt-and-pepper noise) are leveraged to augment SE data to yield TE. To gain a clearer understanding of traditional data augmentation methods, we show the effects of these methods on images from the Satellite Image Classification (SIC) dataset \cite{satelliteClassficationDataset} in Figs. \ref{rotation_data}$\sim$\ref{salt_and_pepper_data}. It is noteworthy that Fig. \ref{original_data} is the original data from the SIC dataset. With respect to the backbone network, we employ Swin-T (belongs to Swin Transformer) \cite{liu2021swin} and ResNet10 (a type of ResNets \cite{he2016deep}) for simplicity. 

\par
\textbf{Baselines.} \quad We compare the proposed GK-FedDKD method with several state-of-the-art (SOTA) benchmarks: \textbf{FedExP} \cite{jhunjhunwala2023fedexp}, \textbf{MOON} \cite{li2021model}, \textbf{FedAU} \cite{wang2023lightweight}, \textbf{FedProto} \cite{tan2022fedproto}, \textbf{FedProx} \cite{li2020federated}, \textbf{FedAS} \cite{yang2024fedas} and \textbf{FedPer} \cite{arivazhagan2019federated}. We also clarify the importance of \textbf{critical components} to the proposed GK-FedDKD method.


\subsection{Comparison in Convergence}
Fig. \ref{convergence_comparison} showcases the convergence curve. As training progresses, the accuracy achieved by the proposed method will converge. Compared with baselines, the proposed method performs best, achieving the highest accuracy at convergence. This may be due to the contributions of each critical component to the proposed method (see Table \ref{sec_4_tab3_Components}) and to the consideration of KD, as applying KD to FL can help cope with heterogeneous data \cite{zou2024cyber}.




\subsection{Comparison in Performance}
Table~\ref{sec_4_tab1_accuracy_AA_MFS_Swin} reports accuracy, average accuracy (AA), and macro-F1 score (MFS) to evaluate the effectiveness of our method. As shown, the proposed GK-FedDKD consistently delivers substantial improvements across all three metrics compared with the baselines. For example, on the EuroSAT dataset with Swin-T as the client backbone, GK-FedDKD surpasses the second-best method by $7.17\%$ in accuracy, $7.42\%$ in AA, and $7.59\%$ in MFS. Similar performance gains are observed across the other datasets and with the ResNet10 backbone. Notably, FedProx yields extremely low scores on EuroSAT when using ResNet10. Therefore, its results are omitted from Table~\ref{sec_4_tab1_accuracy_AA_MFS_Swin}. We attribute this phenomenon to the fact that both KD and the prototype aggregation policy are incorporated into the proposed approach, where KD is known as a technique for handling data heterogeneity, and the prototype aggregation mechanism can be leveraged to support heterogeneous clients \cite{zou2024cyber}. It is noteworthy that although FedProto also adopts a prototype aggregation strategy, its performance (i.e., accuracy, average accuracy, and macro-F1 score) remains lower than that of the proposed method. Hence, we can infer that the joint adoption of KD and the prototype aggregation strategy is more conducive to performance improvement. For simplicity, Swin-T is adopted as the default backbone in the remainder of the experiments. 


\begin{table*}[t!]
\renewcommand{\thetable}{\Roman{table}}
\setlength{\extrarowheight}{1 pt}
\caption{Importance of critical components for the proposed method (MPG: multi-prototype generation, LL\_M: linear layer-based module, GIA: global information alignment).}
\begin{center}
\begin{tabular}{|m{1.9cm}|m{1.9cm}|m{1.9cm}|m{1.9cm}|m{2.8cm}|m{1.9cm}|m{1.9cm}|} \hline
	\multirowcell{2}{\bfseries Dataset} & \multirowcell{2}{\bfseries MPG} & \multirowcell{2}{\bfseries LL\_M} & \multirowcell{2}{\bfseries GIA} & \multirowcell{2}{\bfseries TE Generation} & \multicolumn{2}{c|}{\bfseries Swin-T}\\
	\cline{6-7}
	  &  & & & & \centering Accuracy & \hfil \textbf{$\bigtriangleup$} \\
	\hline
	\multirowcell{5}{EuroSAT}
	& \hfil \faTimes & \hfil \faCheck & \hfil \faCheck & \hfil \faCheck & \hfil $0.8977$  & \hfil $-0.0026$\\ \cline{2-7}
	& \hfil \faCheck & \hfil \faTimes & \hfil \faCheck & \hfil \faCheck & \hfil $0.8958$ & \hfil $-0.0045$  \\ \cline{2-7}
	& \hfil \faCheck & \hfil \faCheck  & \hfil \faTimes  & \hfil \faCheck  & \hfil $0.8974$ & \hfil  $-0.0029$ \\ \cline{2-7}
    & \hfil \faCheck & \hfil \faCheck & \hfil \faCheck & \hfil \faTimes & \hfil $0.8961$ & \hfil $-0.0042$  \\ \cline{2-7}
    & \hfil \faCheck & \hfil \faCheck & \hfil \faCheck & \hfil \faCheck & \hfil $0.9003$ & \hfil $-$  \\ \cline{2-7}
	\hline \hline
    \multirowcell{5}{SIC}
	& \hfil \faTimes & \hfil \faCheck & \hfil \faCheck & \hfil \faCheck & \hfil $0.8908$  & \hfil $-0.0324$\\ \cline{2-7}
	& \hfil \faCheck & \hfil \faTimes  & \hfil \faCheck & \hfil \faCheck  & \hfil $0.9166$  & \hfil $-0.0066$  \\ \cline{2-7}
	& \hfil \faCheck & \hfil \faCheck & \hfil \faTimes  & \hfil \faCheck & \hfil $0.9070$ & \hfil  $-0.0162$ \\ \cline{2-7}
    & \hfil \faCheck & \hfil \faCheck & \hfil \faCheck & \hfil \faTimes & \hfil $0.9028$ & \hfil $-0.0204$  \\ \cline{2-7}
    & \hfil \faCheck & \hfil \faCheck & \hfil \faCheck & \hfil \faCheck & \hfil $0.9232$ & \hfil  $-$ \\ \cline{2-7}
	\hline 
\end{tabular}
\end{center}
\label{sec_4_tab3_Components}
\end{table*}

\par
In Table \ref{sec_4_tab2_Error_Comparison}, we leverage various metrics, i.e., error rate and mean absolute error (MAE), to showcase the performance of the proposed method by comparing it with FedExP, MOON, FedProto, and FedAU. The results in Table \ref{sec_4_tab2_Error_Comparison} show that the proposed method achieves a notable reduction in error rate and MAE across the four datasets. Hence, we can infer that the proposed method surpasses the baselines.
\begin{table}[t!]
\renewcommand{\thetable}{\Roman{table}}
\setlength{\extrarowheight}{1 pt}
\caption{Accuracy obtained by FedAS, FedPer and the proposed GK-FedDKD approach when considering various learning rates for the SAT6 dataset. The best value is highlighted in bold, the second best is shown in blue, and the enhancement is offered in red.}
\begin{center}
\begin{tabular}{|m{1.3cm}|m{1.3cm}|m{1.3cm}|m{1.3cm}|m{1.3cm}|}\hline
	 \multirowcell{2}{\bfseries LR}
	& \multicolumn{4}{c|}{\bfseries Accuracy} \\%
	\cline{2-5}
	  & \thead{\bfseries FedAS} & \thead{\bfseries FedPer} & \thead{\bfseries Ours} & \hfil \color{red}{\textbf{$\bigtriangleup$}} \\%
	\hline
	  \thead{0.01} & \hfil $0.8445$  & \hfil $\color{blue}{0.8490}$ & $\hfil \textbf{0.9650}$ & \hfil $\color{red}{+0.116}$\\%
	\hline
	  \thead{0.001} & \hfil $0.7415$ & \hfil $\color{blue}{0.7915}$ & $\hfil \textbf{0.9485}$ & \hfil $\color{red}{+0.157}$ \\%
	\hline
	  \thead{0.0001} & \hfil $0.5155$ & \hfil $\color{blue}{0.6860}$ &  \hfil $\textbf{0.9175}$ & \hfil $\color{red}{+0.2315}$ \\%
	\hline
\end{tabular}
\end{center}
\label{sec_4_tab4_Learning_Rate}
\end{table}

\begin{figure} [t!]
	\centering
	\includegraphics[scale = 0.3]{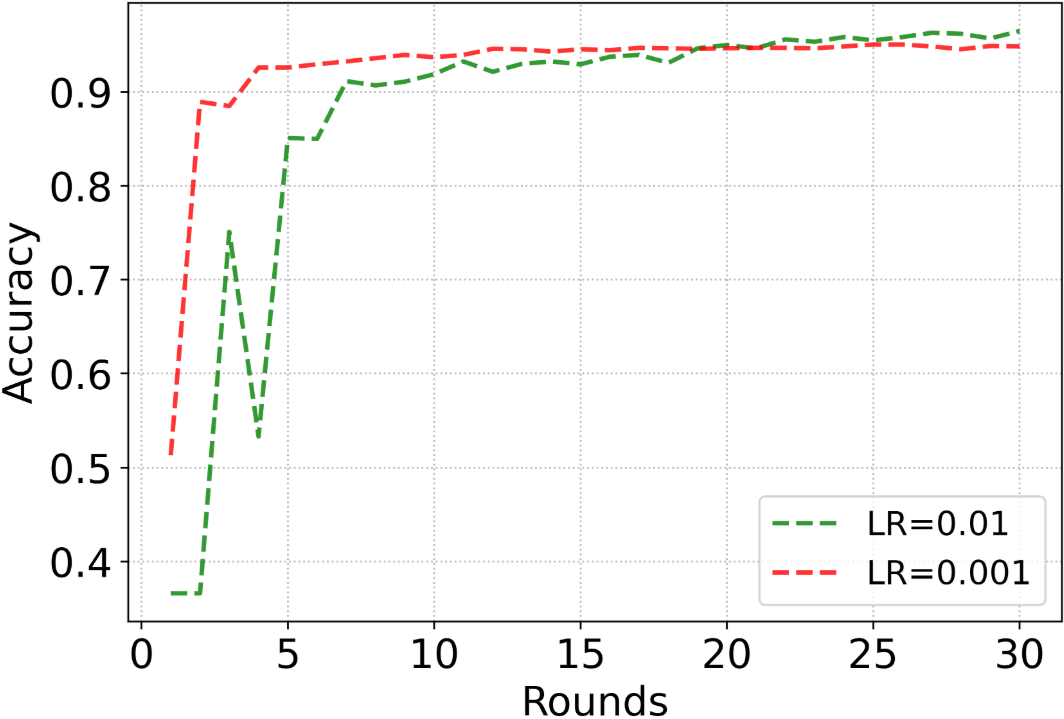}
	\captionsetup{font=small}
	\caption{The convergence curve was obtained by using the proposed GK-FedDKD approach over SAT6 dataset under the consideration of $LR=0.01$ and $LR=0.001$.}
	\label{fig_convergence_curve_LR}
    \vspace{-0.3cm}
\end{figure}

\subsection{Ablation Study}
The impact of essential components and the learning rate to the proposed method will be introduced in this part.
\par
\textbf{Components of the proposed GK-FedDKD.} \quad In Table \ref{sec_4_tab3_Components}, the importance of critical components for the proposed method is analyzed. Concretely, we detect the importance by discarding each component, i.e., multi-prototype generation (MPG), linear layer-based module (LL\_M), global information alignment (GIA), and TE generation. For convenience of comparison, we treat the accuracy obtained by the proposed method as a benchmark. From Table \ref{sec_4_tab3_Components}, we see that the benchmark accuracies for the EuroSAT and SIC datasets are $0.9003$ and $0.9232$, respectively, the highest values. Additionally, abolishing one of the aforementioned components will reduce accuracy. Accordingly, we can conclude that each component is crucial to the proposed method.

\begin{table}[t!]
	\renewcommand{\thetable}{\Roman{table}}
	\setlength{\extrarowheight}{1 pt}
	\caption{Accuracy achieved by the proposed method and the baselines under the consideration of various numbers of clients (i.e., $\{10, 20, 30, 50\}$), where SAT6 dataset is adopted. The optimal values are shown in bold, and the second-best is marked in blue. The improvements between them are provided in red.}
	\begin{center}
		\begin{tabular}{|p{0.115\textwidth}|>{\centering}p{0.06\textwidth}|>{\centering}p{0.06\textwidth}|>{\centering\arraybackslash}p{0.06\textwidth}|p{0.06\textwidth}|
			}
			\hline	
			\centering Method & \#$10$ & \#$20$ & \#$30$ & \hfil \#$50$\\
			\hline
			\centering FedExP \cite{jhunjhunwala2023fedexp} & $0.8875$ & $0.8387$ & $\color{blue}{0.9045}$  & $\color{blue}{0.9270}$ \\ \hline
			\centering MOON \cite{li2021model} & $0.7335$ & $0.7521$ & $0.7536$ & $0.5975$ \\ \hline
			\centering FedAU \cite{wang2023lightweight} & $\color{blue}{0.9240}$ & $0.8280$ & $0.6433$  & $0.5782$ \\ \hline
			\centering FedProto \cite{tan2022fedproto}  & $0.8740$ & $\color{blue}{0.8857}$ & $0.8567$ & $0.7560$ \\ \hline
			\centering FedProx \cite{li2020federated} & $0.4075$ & $0.5726$ & $0.7034$ & $0.6870$ \\ \hline
			\centering FedAS \cite{yang2024fedas}  & $0.7415$ & $0.7746$ & $0.5960$ & $0.5517$ \\ \hline
			\centering FedPer \cite{arivazhagan2019federated} & $0.7915$ & $0.8323$ & $0.6948$ & $0.6578$ \\ \hline
			\centering \textbf{Ours} & $\textbf{0.9485}$  & $\textbf{0.9460}$ & $\textbf{0.9176}$ & $\textbf{0.9610}$\\ \hline 
			\hfil $\color{red}{\bigtriangleup}$ & $\color{red}{+0.0245}$  & $\color{red}{+0.0603}$ & $\color{red}{+0.0131}$ & $\color{red}{+0.034}$ \\ \hline 
		\end{tabular}
	\end{center}
	\label{supplementary_num_clients}
    \vspace{-0.2cm}
\end{table}

\par
\textbf{Learning Rate.} \quad In Table \ref{sec_4_tab4_Learning_Rate}, we attempt to explore the influence of the learning rate for the proposed method. To be specific, we observe that with a learning rate of $0.01$, the accuracy reaches its best value. However, in this work, we set the default learning rate to $0.001$. The reason is that, for the proposed method, the convergence curve (as shown in Fig. \ref{fig_convergence_curve_LR}) exhibits greater volatility in the initial stage when LR=0.01. Second, when LR=0.001, the proposed method converges faster. It is worth noting that, compared with FedAS and FedPer, the proposed method achieves the highest accuracy regardless of the learning rate. Thus, Table \ref{sec_4_tab4_Learning_Rate} also indicates that the performance of the proposed method is superior to that of FedAS and FedPer.

\par

\textbf{Number of Clients.} \quad
In Table \ref{supplementary_num_clients}, the effectiveness of the proposed method is further explored, with various numbers of clients (i.e., ${10, 20, 30, 50}$) and a Dirichlet distribution parameter of $0.9$ \cite{zou2024cyber}. Besides, the SAT6 dataset is employed as a case study for simplicity. Compared with the baselines, the proposed method achieves the highest accuracy with values of $0.9485$, $0.9460$, $0.9176$, and $0.9610$, respectively, for the number of clients: $10$, $20$, $30$, and $50$. To be more specific, when the number of clients is $10$, $20$, $30$ and $50$, the accuracy obtained by the proposed method is around $2.65\%$, $6.81\%$, $1.45\%$ and $3.67\%$ greater than the second best, respectively. Notice that it is unreasonable to compare accuracy across different numbers of clients. This is because we employ a random client selection method, which may result in some clients not participating in training. Since the proposed method achieves the best results in Table \ref{supplementary_num_clients}, we can infer that it is superior to the baselines.

\par

\begin{table*}[t!]
	\renewcommand{\thetable}{\Roman{table}}
	\setlength{\extrarowheight}{1 pt}
	\caption{Performance comparison among the proposed GK-FedDKD and other baselines (AA: Average Accuracy; MAS: Macro-F1 Score) over SAT6 dataset. The largest values are depicted in bold, the second-best is highlighted in blue, and the improvement between the best and the second-best is shown in red.}
	\begin{center}
		\begin{tabular}{p{0.12\textwidth}>{\centering}p{0.12\textwidth}>{\centering\arraybackslash}p{0.1\textwidth}>{\centering}p{0.12\textwidth}|>{\centering\arraybackslash}p{0.1\textwidth}>{\centering\arraybackslash}p{0.08\textwidth}>{\centering\arraybackslash}p{0.1\textwidth}}
			\hline \hline
			\centering \multirow{2}{*}{Method} &\multicolumn{3}{c|}{$Dir=0.5$} & \multicolumn{3}{c}{$Dir=50$} \\ \cline{2-7}
			& Accuracy & AA & MFS & Accuracy & AA & MFS \\ \hline
			\hline
			\centering	FedExP \cite{jhunjhunwala2023fedexp}  & $0.8665$ & $0.8397$ & $\color{blue}{0.7564}$  & $\color{blue}{0.9325}$ & $\color{blue}{0.9346}$ & $\color{blue}{0.8261}$ \\
			\centering	MOON \cite{li2021model} & $0.7595$  & $0.7299$ & $0.4856$  & $0.8945$ & $0.8940$  & $0.7222$ \\ 
			\centering	FedAU \cite{wang2023lightweight} & $0.88$ & $0.8419$ & $0.6836$ & $0.9320$ & $0.9321$ & $0.7577$  \\
			\centering	FedProto \cite{tan2022fedproto} & $\color{blue}{0.8855}$ & $\color{blue}{0.8610}$ & $0.7417$ & $0.9265$ & $0.9263$ & $0.8252$ \\  
			\centering	FedProx \cite{li2020federated}  & $0.5535$ & $0.5591$ & $0.3372$ & $0.3490$ & $0.3491$ & $0.1745$ \\
			\centering	FedAS \cite{yang2024fedas} & $0.7435$ & $0.7218$ & $0.5472$ & $0.7950$ & $0.7945$ & $0.6269$ \\  
			\centering	FedPer \cite{arivazhagan2019federated} & $0.8385$ & $0.8221$ & $0.6584$ & $0.9205$ & $0.9208$ & $0.7980$ \\ \hline 
			\centering	\textbf{Ours} & $\textbf{0.9310}$ & $\textbf{0.9099}$ & $\textbf{0.8423}$ & $\textbf{0.9580}$ & $\textbf{0.9578}$ & $\textbf{0.9109}$ \\ \hline 
			\centering	$\color{red}{\bigtriangleup}$ & $\color{red}{+0.0455}$ & $\color{red}{+0.0489}$ & $\color{red}{+0.0859}$ & $\color{red}{+0.0255}$ & $\color{red}{+0.0232}$ & $\color{red}{+0.0848}$ \\ \hline \hline
		\end{tabular}
	\end{center}
	\label{supp_dir}
\end{table*}
\textbf{Various Parameters for Dirichlet Distribution.} \quad 
To exhibit the performance of the proposed GK-FedDKD approach, in Table \ref{supp_dir}, we compare the proposed method with several baselines by considering using various parameters for the Dirichlet distribution (i.e., $Dir=0.5$ and $Dir=50$) to prepare the heterogeneous data. For convenience, the SAT6 dataset is adopted. It is worth noting that the parameter of the Dirichlet distribution determines the degree of data heterogeneity, where lower values produce more imbalanced class distributions among clients. In Table \ref{supp_dir}, it can be seen that the accuracy, average accuracy (AA), and macro-F1 Score (MFS) achieved by the proposed method are the highest. Particularly, the proposed method outperforms the second best by $4.55\%$ in accuracy, $4.89\%$ in AA, and $8.59\%$ in MFS, respectively, when $Dir=0.5$. When $Dir=50$, the improvements become $2.55\%$ in accuracy, $2.32\%$ in AA, and $8.48\%$ in MFS. Accordingly, it can be inferred that the proposed method is optimal compared with the baselines.

\subsection{Visualization}
\par

\begin{figure}[t!]
	\centering
	\begin{subfigure}{0.42\textwidth} 
		\includegraphics[width=0.96\linewidth]{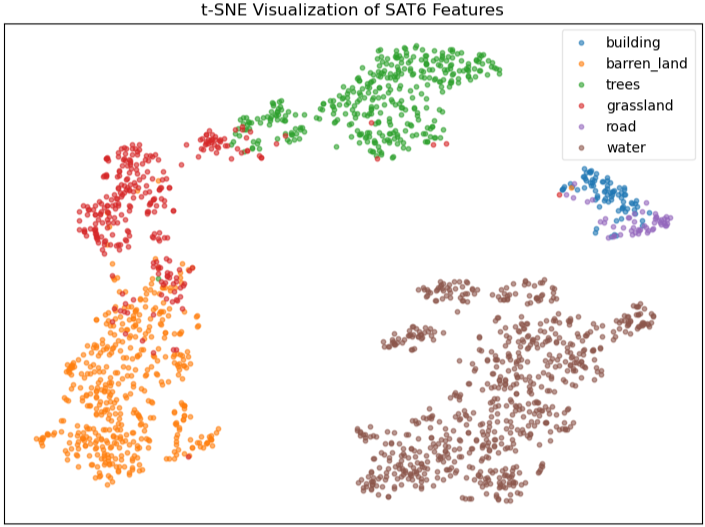}
		\captionsetup{font=small} 
		\caption{Global model trained by using SAT6 dataset (Dir=0.5)}
		\label{SAT6_t_SNE_Dir0_5}
	\end{subfigure} \\
	\begin{subfigure}{0.42\textwidth} 
		\includegraphics[width=0.96\linewidth]{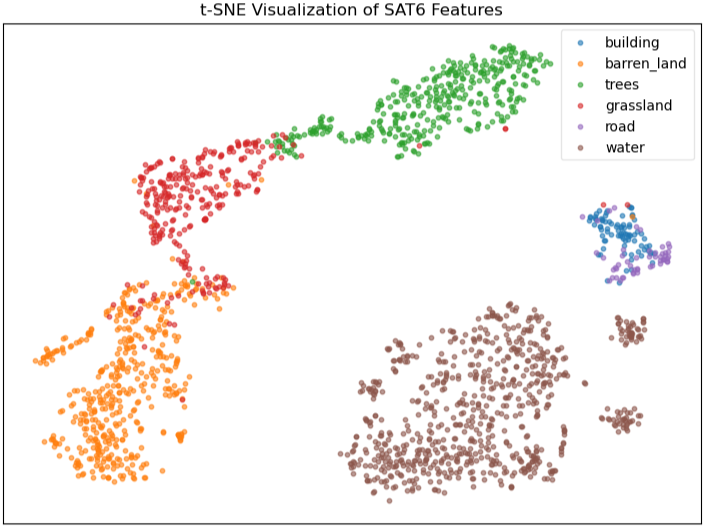}
		\captionsetup{font=small} 
		\caption{Global model trained by using SAT6 dataset (Dir=50)}
		\label{SAT6_t_SNE_Dir50}
	\end{subfigure}	
	\captionsetup{font=small} 
	\caption {Local embedding visualization for the test part of SAT6 dataset. The corresponding local embedding is generated by using the global model trained by the proposed method in round $30$.}
	\label{tSNE_Visualization}
\end{figure}

\begin{figure*}[t!]
	\centering
	\begin{subfigure}{0.47\textwidth} 
		\includegraphics[width=1.\linewidth]{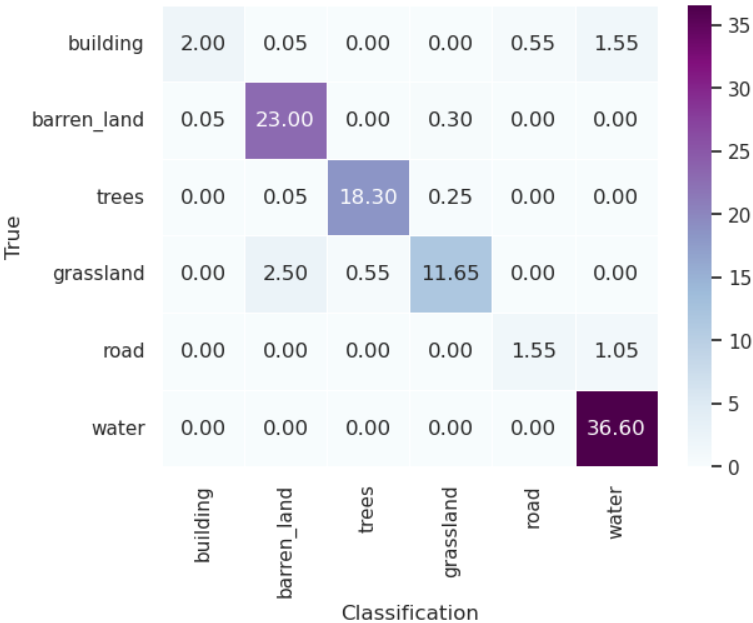}
		\captionsetup{font=small} 
		\caption{SAT6 dataset (Dir=0.5)}
		\label{SAT6_CM_0_5}
	\end{subfigure} \hfill
	\begin{subfigure}{0.47\textwidth} 
		\includegraphics[width=1.\linewidth]{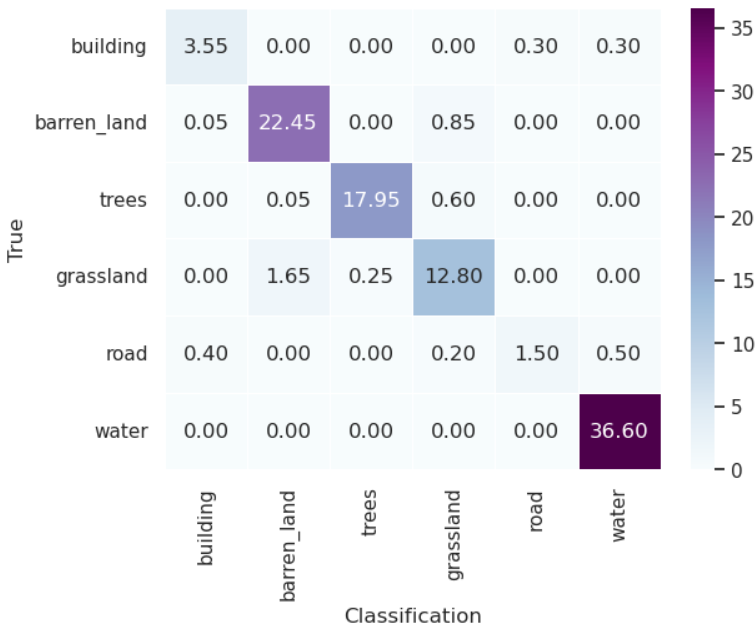}
		\captionsetup{font=small} 
		\caption{SAT6 dataset (Dir=0.9)}
		\label{SAT6_CM_0_9}
	\end{subfigure} \\
	\begin{subfigure}{0.47\textwidth} 
		\includegraphics[width=1.\linewidth]{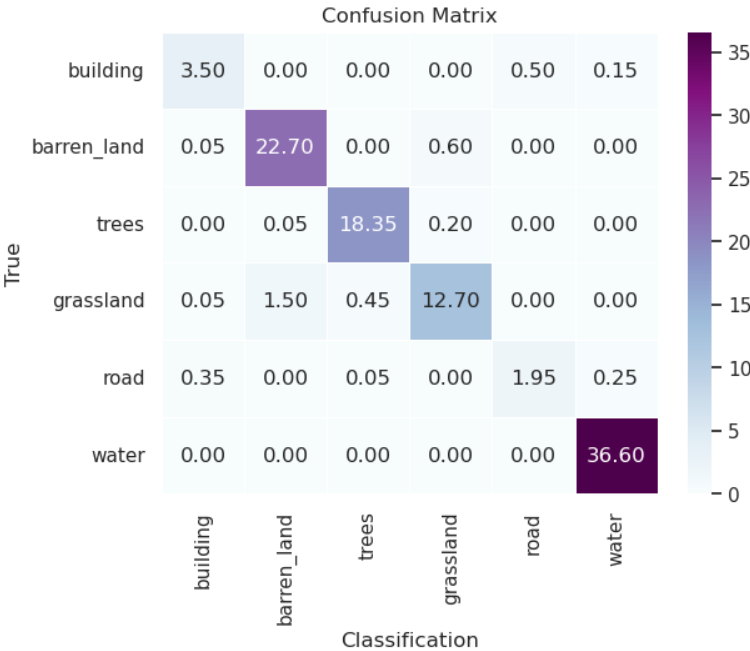}
		\captionsetup{font=small} 
		\caption{SAT6 dataset (Dir=50)}
		\label{SAT6_CM_50}
	\end{subfigure} \hfill
	\begin{subfigure}{0.47\textwidth} 
		\includegraphics[width=1.\linewidth]{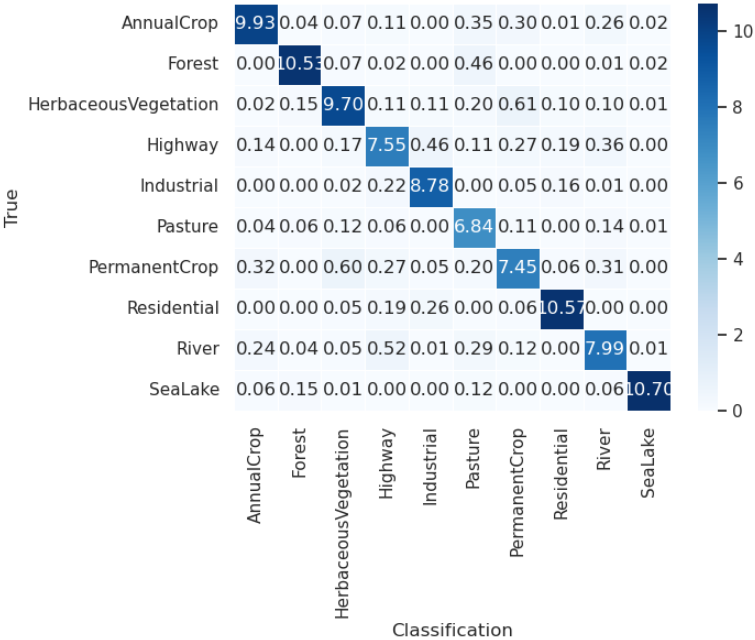}
		\captionsetup{font=small} 
		\caption{EuroSAT (Dir=0.9)}
		\label{EuroSAT_CM}
	\end{subfigure} \\
	\begin{subfigure}{0.47\textwidth} 
		\includegraphics[width=1.\linewidth]{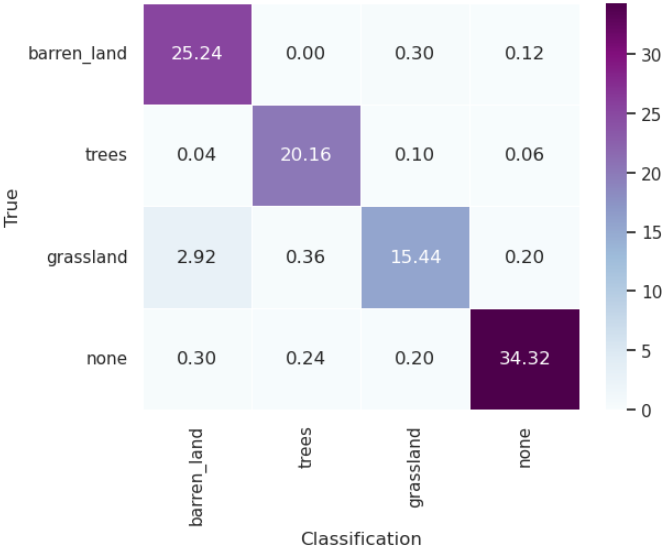}
		\captionsetup{font=small} 
		\caption{SAT4 dataset (Dir=0.9)}
		\label{SAT4_CM}
	\end{subfigure} \hfill
	\begin{subfigure}{0.47\textwidth} 
		\includegraphics[width=1.\linewidth]{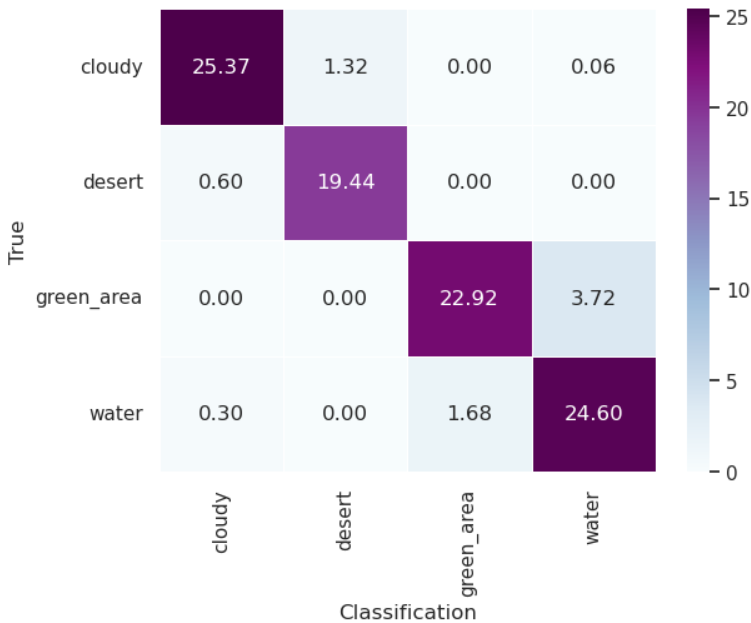}
		\captionsetup{font=small} 
		\caption{SIC dataset (Dir=0.9)}
		\label{SIC_CM}
	\end{subfigure}	
	\captionsetup{font=small} 
	\caption {Confusion matrix achieved by leveraging the proposed GK-FedDKD approach.}
	\label{Confusion_Matrix_proposed}
\end{figure*}

In Fig. \ref{tSNE_Visualization}, we use t-SNE to visualize the image features (local embeddings) obtained by the proposed method with $Dir=0.5$ and $Dir=50$, using the test set of the SAT6 dataset for simplicity. Specifically, the image features are generated using the global model trained in the $30^{th}$ round on data distributed according to Dirichlet distributions with parameters $0.5$ and $50$. As we can see, most of the feature representations (local embeddings) obtained using the trained two global models can be easily distinguished. Therefore, the proposed method is effective for training the global model under $Dir=0.5$ and $Dir=50$.

\par
In Fig.~\ref{Confusion_Matrix_proposed}, to acquire more profound insights into the performance of the proposed method, we visualize the results gained by the proposed GK-FedDKD approach with a confusion matrix in percentage, where the values on the diagonal represent the accuracy of each category. Particularly, considering 0.9 as the parameter of the Dirichlet distribution, we show the confusion matrix for the considered four datasets, i.e., the SAT6 dataset (Fig. \ref{SAT6_CM_0_9}), EuroSAT dataset (Fig. \ref{EuroSAT_CM}), SAT4 dataset (Fig. \ref{SAT4_CM}), and SIC
dataset (Fig. \ref{SIC_CM}). In addition to these, we additionally yield the confusion matrix for the SAT6 dataset distributed by considering $Dir=0.5$ and $Dir=50$. From these confusion matrices, it can be seen that the proposed method is applicable across various datasets and different Dirichlet distribution parameters.   

\section{Conclusion}
\label{sec_5_conclusion}
In this work, we have proposed a geometric knowledge-guided federated dual knowledge distillation (GK-FedDKD) method to address data heterogeneity in remote sensing satellite imagery. This method involves two knowledge distillation techniques, a multi-prototype generation strategy, local data augmentation with geometric knowledge, and a novel linear layer-based module. Experiments demonstrate that the proposed GK-FedDKD outperforms SOTA methods across four datasets. For instance, the accuracy for the SAT6 dataset when considering 0.5 as the parameter of the Dirichlet distribution is $1.07\times$, $1.23\times$, $1.06\times$, $1.05\times$, $1.68\times$, $1.25\times$, and $1.11\times$ greater than FedExP, MOON, FedAU, FedProto, FedProx, FedAS, and FedPer, respectively.


\section*{Acknowledgments}
We would like to express our sincere gratitude to Prof. Dusit Niyato from Nanyang Technological University, Singapore (e-mail: dniyato@ntu.edu.sg) for providing valuable comments to improve the quality of this paper.


\bibliographystyle{IEEEtran}

\bibliography{Ref}


\newpage

 




\vfill

\end{document}